\documentclass[conference]{IEEEtran}
\usepackage{times}
\usepackage{booktabs}
\usepackage{multirow}
\usepackage{graphicx}
\usepackage{subcaption}
\usepackage{amsmath}
\usepackage{amssymb}
\usepackage{algorithm}
\usepackage{algpseudocode}
\usepackage{amsmath}
\usepackage{amssymb}
\usepackage{booktabs}
\usepackage{arydshln}
\usepackage[margin=15mm]{geometry}
\usepackage{amsmath}

\usepackage[numbers]{natbib}
\usepackage{multicol}
\usepackage[bookmarks=true]{hyperref}

\usepackage{eso-pic}

\IEEEoverridecommandlockouts

\begin{document}

\AddToShipoutPictureBG*{%
  \AtPageUpperLeft{%
    \hspace{0.5\paperwidth}%
    \raisebox{-1.1cm}{%
      \makebox[0pt][c]{%
        \parbox{\textwidth}{%
          \centering
          \normalsize
          \textit{This paper has been accepted for publication in Robotics: Science and Systems (RSS) 2025.}
        }%
      }%
    }%
  }%
}
\title{Doppler Correspondence: Non-Iterative Scan Matching With Doppler Velocity-Based Correspondence}

\author{
    Jiwoo Kim$^{*, \dagger, 1}$, 
    Geunsik Bae$^{*, 2}$, 
    Changseung Kim$^{1}$, 
    Jinwoo Lee$^{1}$, 
    Woojae Shin$^{2}$, 
    and Hyondong Oh$^{\dagger\dagger, 2}$%
    \thanks{$^{*}$Equal contributions. $^{\dagger}$Project lead. $^{\dagger\dagger}$Corresponding author.}%
    \thanks{$^{1}$Department of Mechanical Engineering, Ulsan National Institute of Science and Technology, Republic of Korea.}%
    \thanks{$^{2}$Department of Mechanical Engineering, Korea Advanced Institute of Science and Technology, Republic of Korea.}%
    \thanks{Emails: \{tars0523, pon02124, jinwoolee2021\}@unist.ac.kr; \{baegs94, oj7987, h.oh\}@kaist.ac.kr}%
}

\maketitle
\begin{abstract}
Achieving successful scan matching is essential for LiDAR odometry. However, in challenging environments with adverse weather conditions or repetitive geometric patterns,  LiDAR odometry performance is degraded due to incorrect scan matching. Recently, the emergence of frequency-modulated continuous wave 4D LiDAR and 4D radar technologies have provided the potential to address these unfavorable conditions. The term 4D refers to point cloud data characterized by range, azimuth, and elevation along with Doppler velocity. Although 4D data is available, most scan matching methods for 4D LiDAR and 4D radar still establish correspondence by repeatedly identifying the closest points between consecutive scans, overlooking the Doppler information. This paper introduces, for the first time, a simple Doppler velocity-based correspondence---\textit{Doppler Correspondence}---that is invariant to translation and small rotation of the sensor, with its geometric and kinematic foundations. Extensive experiments demonstrate that the proposed method enables the direct matching of consecutive point clouds without an iterative process, making it computationally efficient. Additionally, it provides a more robust correspondence estimation in environments with repetitive geometric patterns. The implementation of our proposed method is publicly available at \url{https://github.com/Tars0523/Doppler_Correspondence}.

\end{abstract}

\IEEEpeerreviewmaketitle

\section{Introduction}
State estimation is a fundamental component in autonomous systems, enabling robots to achieve tasks accurately and safely~\cite{doer2020ekf, hexsel2022dicp, xu2024modeling}. Accurate odometry plays a critical role in reliable state estimation by tracking the robot's movement and position. Therefore, LiDAR odometry has been extensively researched to achieve high accuracy and real-time performance, leveraging its superior precision and detailed environmental sensing capabilities. However, scan matching, which is essential for LiDAR odometry, is degraded in adverse weather conditions such as rain, smoke, and fog~\cite{nissov2024degradation}. Additionally, most scan matching methods, which depend on the geometry of the environment, often fail to accurately estimate sensor motion in repetitive geometric structures such as tunnels and highways~\cite{hexsel2022dicp}.

Recently, the frequency-modulated continuous wave (FMCW) 4D LiDAR and 4D radar have shown great potential in alleviating these limitations. Unlike conventional LiDAR, FMCW-based 4D ranging sensors are more resilient to weather conditions~\cite{lee2023frequency,nissov2024degradation}. Moreover, they provide 4D point cloud information (range, azimuth, elevation, and Doppler velocity), which has potential~\cite{hexsel2022dicp, nissov2024degradation} for robust scan matching in scenarios with repetitive geometric structures.

Existing approaches related to correspondence estimation for 4D point cloud scan matching either rely on heuristic utilization of radar cross section (RCS)~\cite{huang2024less} or leverage the uncertainty inherent in point clouds derived from 3D information~\cite{xu2024modeling}. Although FMCW-based 4D ranging sensors provide Doppler velocity but still the correspondence estimation methods typically rely solely on 3D information~\cite{herraez2024radar, hexsel2022dicp, li20234d, michalczyk2022tightly, michalczyk2023multi, zhang20234dradarslam}.

\begin{figure}[t!]
	\centering
	\includegraphics[width=0.90\linewidth]{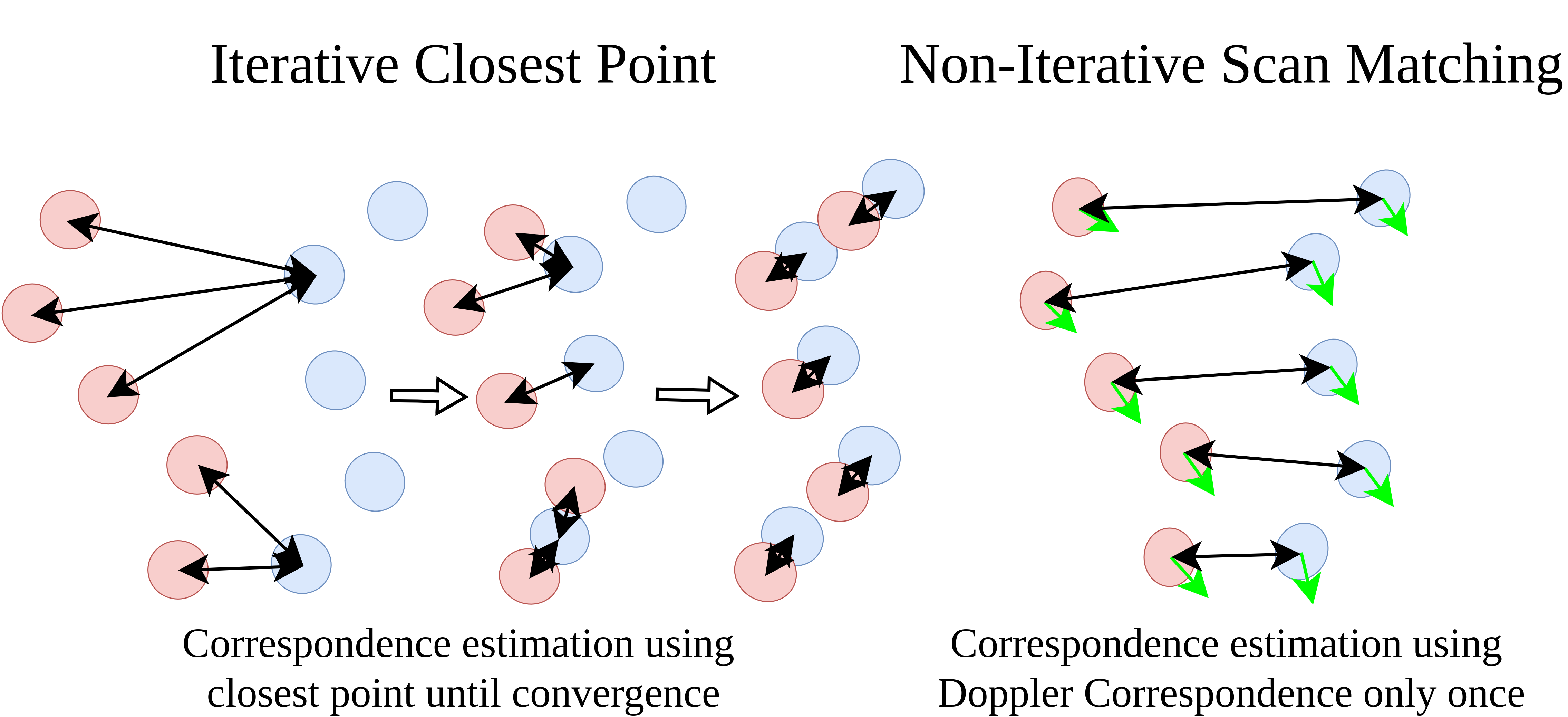}
	\caption{The red circle denotes source points, the blue circle represents target points, and the green arrow indicates the Doppler velocity of each point. While the ICP method requires multiple iterations to refine correspondences, the proposed method utilizes the direct scan matching approach based on Doppler Correspondence.}
	\label{fig: non iter}
	\vspace{-1mm}
\end{figure}
\begin{figure}[t!]
	\centering
	\includegraphics[width=0.85\linewidth]{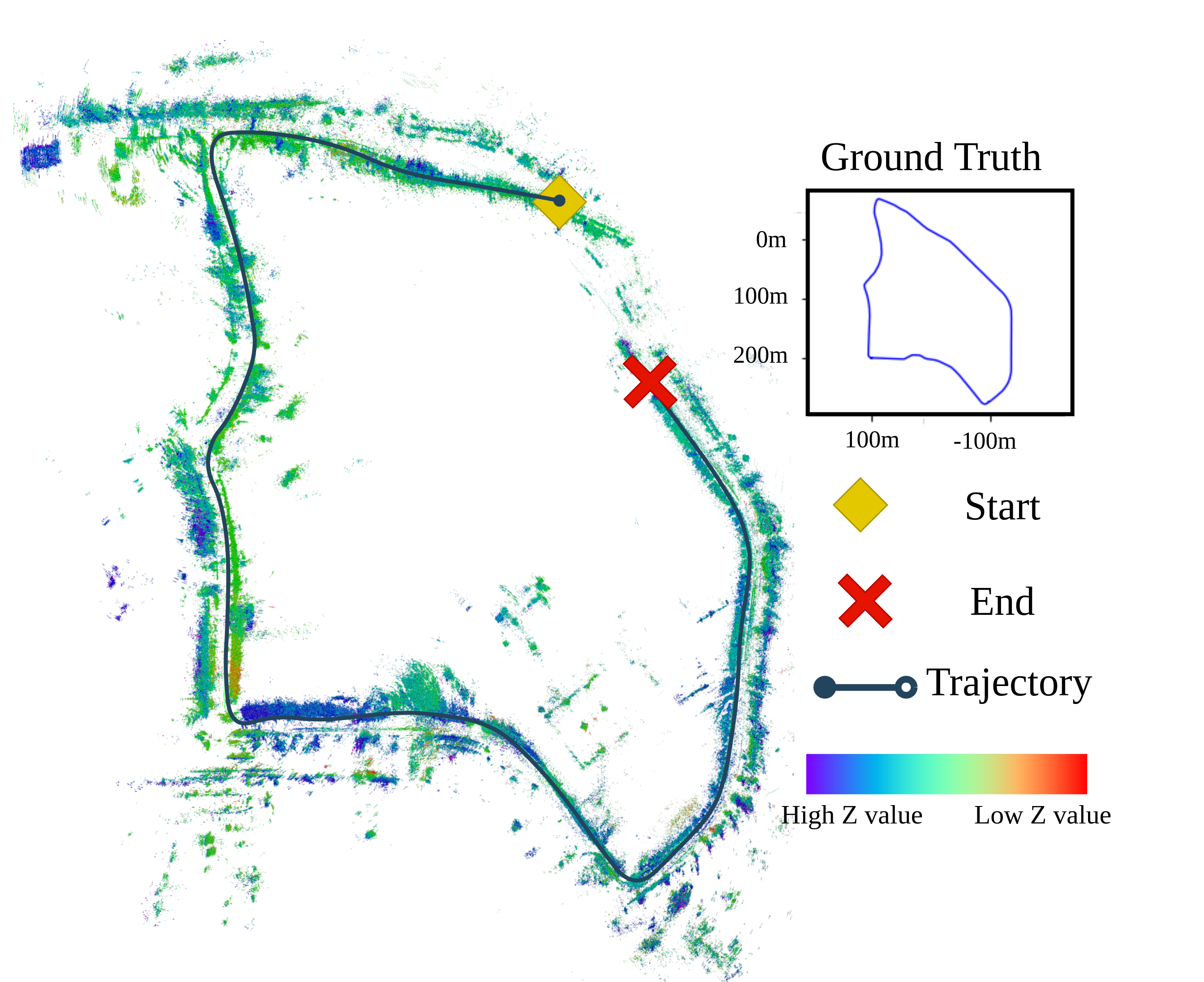}
	\caption{The result of the proposed non-iterative scan matching method using the \textit{nyl} trajectory in the NTU4DRadLM dataset~\cite{zhang2023ntu4dradlm}.}
	\label{fig: non_iter_nyl}
	\vspace{-3mm}
\end{figure}

In this paper, for the first time, --- \textbf{\textit{Doppler Correspondence}}---a simple correspondence that leverages Doppler velocity is introduced. This correspondence is based on the range and Doppler velocity of each point. Unlike the iterative closest point (ICP), scan-matching utilizing Doppler Correspondence directly matches point clouds without an iterative process, as illustrated in Fig.~\ref{fig: non iter}, while preserving a reasonable level of odometry accuracy, as shown in Fig.~\ref{fig: non_iter_nyl}. The key contributions of this work are summarized as follows:

\begin{itemize}
\item A simple and novel correspondence utilizing Doppler velocity is proposed for the first time. Geometric and kinematic foundations of this correspondence are derived; 
\item Doppler Correspondence enables a non-iterative scan matching. Consequently, it significantly reduces time consumption compared with traditional ICP methods;
\item Since the proposed correspondence does not depend on geometric cues alone, it remains effective even in the presence of repetitive geometric structures; and
\item Scan matching based on Doppler Correspondence achieves performance comparable to ICP methods across diverse scenarios and sensor configurations with less computation time.
\end{itemize}
\vspace{-3mm}
\section{Related Work}
\subsection{4D Radar Odometry}
4D radar odometry and simultaneous localization and mapping (SLAM) methods have gained significant attention due to their robustness in challenging environments, such as visually degraded or bad weather conditions. Early work by \citet{doer2020ekf,doer2020radar} demonstrated the feasibility of fusing 4D radar data with inertial measurements to achieve 4D radar-inertial odometry (RIO). Using extended kalman filter (EKF)-based approaches, they highlighted improvements in pose estimation through techniques such as RANSAC-based ego-velocity estimation, barometric height fusion, and online 4D radar extrinsic calibration, which removed the need for tedious pre-calibration processes. Building on this foundation, \citet{michalczyk2022tightly, michalczyk2023multi} extended RIO methods, incorporating persistent landmark detection and multi-state estimation frameworks to improve performance.

More recently, as 4D radar point clouds become more dense, researchers have focused on optimization-based approaches. Unlike earlier Kalman filter-based methods, which rely on sequential state estimation, optimization-based methods aim to solve the sub-trajectory by considering the global consistency of sensor data. \citet{li20234d} developed a pose graph optimization-based 4D radar SLAM framework that utilizes ego-velocity pre-integration factors to improve robustness in noisy 4D radar data. \citet{huang2024less} introduced an enhanced RIO that integrates Doppler velocity and RCS information to filter noisy point cloud, adhering to the \textit{less is more} principle. Additionally, \citet{nissov2024degradation} tackled the challenges of LiDAR degeneracy by proposing a LiDAR-4D radar-inertial fusion, leveraging 4D radar's robustness in the fog-filled hallway, while preserving LiDAR accuracy in well-conditioned environments. \citet{herraez2024radar} presented methods for pose estimation and map creation, highlighting the benefits of velocity-aided odometry and map filtering to enhance accuracy. \citet{zhang20234dradarslam} and \citet{xu2024modeling} further advanced 4D radar odometry by modeling point uncertainty in polar coordinates, integrating into data association and motion estimation for improved performance in adverse conditions.

\subsection{4D LiDAR Odometry}
4D LiDAR is an emerging technology that enables Doppler velocity measurements, offering advantages in environments with geometric degeneracy or dynamic objects. Several recent studies have explored the potential of this novel sensing modality for odometry and SLAM tasks. \citet{wu2022picking} introduced the first continuous-time 4D LiDAR-only odometry method leveraging Doppler velocity measurements. Their method employs Gaussian process regression to estimate vehicle trajectories and correct motion distortion caused by scanning, significantly outperforming existing methods. Similarly, \citet{hexsel2022dicp} proposed DICP, which integrates Doppler velocity measurements into the ICP framework for robust point cloud registration. By jointly optimizing Doppler and geometric objective functions, their method demonstrated improved registration accuracy and convergence speed, particularly in featureless environments such as tunnels and hallways. \citet{yoon2023need} and~\citet{lisus2024doppler} developed correspondence-free methods that estimate 6-DOF velocity directly from Doppler information without requiring explicit scan matching and incorporate multiple sensors or complementary modules such as IMUs or gyroscopes for full motion estimation.

\subsection{Summary and Limitations of 4D Radar / LiDAR Odometry}
The 4D information of FMCW 4D LiDAR and 4D radar has leveraged its advantages in three main directions. First, Doppler velocity has been effectively incorporated into optimization frameworks to improve accuracy and robustness in challenging environments, such as repetitive geometric settings, foggy conditions, or a significant number of dynamic objects~\cite{doer2020ekf, doer2020radar, herraez2024radar, hexsel2022dicp, li20234d, michalczyk2022tightly, michalczyk2023multi, nissov2024degradation, wu2022picking, zhang20234dradarslam}. Second, scan matching methods have been developed by refining point correspondences, using RCS information~\cite{huang2024less} or the uncertainty of point clouds~\cite{xu2024modeling}. Third, direct motion estimation methods~\cite{yoon2023need, lisus2024doppler} attempt to recover full 6-DOF motion from Doppler information without relying on point correspondences.

While recent advances have shown promising results, most existing correspondence-based methods still rely solely on 3D geometric information, overlooking the rich Doppler velocity available in 4D LiDAR or radar data. Also, correspondence-free approaches estimate 6-DOF motion directly from Doppler measurements but suffer from limited rotational observability, requiring multiple LiDAR sensors or an additional gyroscope to compensate. In this work, we address these limitations by proposing a novel correspondence formulation that explicitly leverages Doppler information. Unlike prior methods, our approach associates Doppler measurements not only with translational motion but also with rotational motion from a single 4D sensor.

The remaining sections are organized as follows. In Section~\ref{sec: Methodology}, Doppler Correspondence is introduced in detail, including its geometric derivation and the rationale behind its robustness in repetitive geometric environments. Section~\ref{sec: exp} presents the experimental results, demonstrating the effectiveness of the proposed method across various datasets and comparing its performance against point-to-point ICP and DICP. In Section~\ref{sec:Limitation-Condition}, the limitations of the approach are introduced, addressing scenarios where the method may face challenges. In Section~\ref{practical usage}, Doppler Correspondence is integrated with ICP to compare its performance with ICP and DICP, demonstrating its potential applicability. Finally, Section~\ref{sec: con} provides the conclusion, summarizing the key contributions of the work and its potential for the field of 4D ranging sensor-based odometry.
\section{Methodology}
\label{sec: Methodology}
\subsection{Problem Definition}
\label{section2}
ICP is a widely used method for point cloud registration, aiming to estimate the relative transformation between two consecutive scans. Let us consider two consecutive point cloud scans, \(\mathcal{P} = \{p_1, p_2, \ldots, p_N\}\) and \(\mathcal{Q} = \{q_1, q_2, \ldots, q_M\}\), where \(\mathcal{P} \) and \(\mathcal{Q} \) contain \(N\) and \(M\) points, respectively. Assume the sensor undergoes a rotation \(\Delta R \in SO(3)\) and a translation \(\Delta t \in \mathbb{R}^{3}\) between scans. To estimate \(\Delta R\) and \(\Delta t\), the three-step process is applied iteratively as below.
\begin{enumerate}
	\item \textbf{Initialization}: The initial predictions for the rotation \(\Delta \hat{R}\) and translation \(\Delta \hat{t}\) are set to \(I_{3 \times 3}\) and \(\begin{bmatrix} 0 & 0 & 0 \end{bmatrix}^\top\), respectively.
	
	\item \textbf{Correspondence Estimation}: Identify correspondences \(\mathcal{C}\) by determining the closest points between the two point cloud scans.
	\[
	\mathcal{C} = \left\{\, (p, q) \;\middle|\; q = \underset{q \in \mathcal{Q}}{\arg \min} \|p \;-\; q\|_2, \; \forall p \in \mathcal{P} \,\right\},
	\]
	where \(p\text{ and } q \in \mathbb{R}^3\).
	\item \textbf{Optimization}: Compute \(\delta R\) and \(\delta t\) that minimize the following loss function:
	\[
	(\delta R, \delta t) = \underset{\delta R, \delta t}{\arg \min} \sum_{(p, q) \in \mathcal{C}} \|q - (\delta R p \;+\; \delta t)\|_2.
	\]
	\item \textbf{Update}: Update the \(\Delta\hat{R}\), \(\Delta\hat{t}\) and source point cloud \(\mathcal{P}\):
	\[
	\Delta\hat{R} \;\leftarrow \; \delta R \Delta\hat{R} , \quad \Delta\hat{t} \;\leftarrow \; \delta t \;+\; \Delta\hat{t},
	\]
	\[
	p \leftarrow \delta R p \;+\; \delta t,\quad \forall p \in \mathcal{P}. 
	\]
\end{enumerate}
\textbf{Correspondence Estimation}, \textbf{Optimization} and \textbf{Update} are repeated until convergence is achieved, typically defined by a threshold on the change in the loss function \(\|q - (\delta R\; p + \delta t)\|^2\) between iterations.

Mostly, the \textbf{Correspondence Estimation} step solely relies on the spatial information of each point cloud. Relying only on the spatial information in \textbf{Correspondence Estimation} makes registration prone to degradation in environments with repetitive geometric structures~\cite{hexsel2022dicp, nissov2024degradation} or when consecutive scans fail to capture the same spatial points due to its sparse and noisy characteristics~\cite{huang2024less}. To address these limitations, this study proposes Doppler Correspondence for the \textbf{Correspondence Estimation}, which is explained from the following in detail.
\subsection{Geometric Derivation of Doppler Correspondence}
\label{sec:3-B}
\begin{figure}[t]
	\centering
	\includegraphics[width=1.0\linewidth]{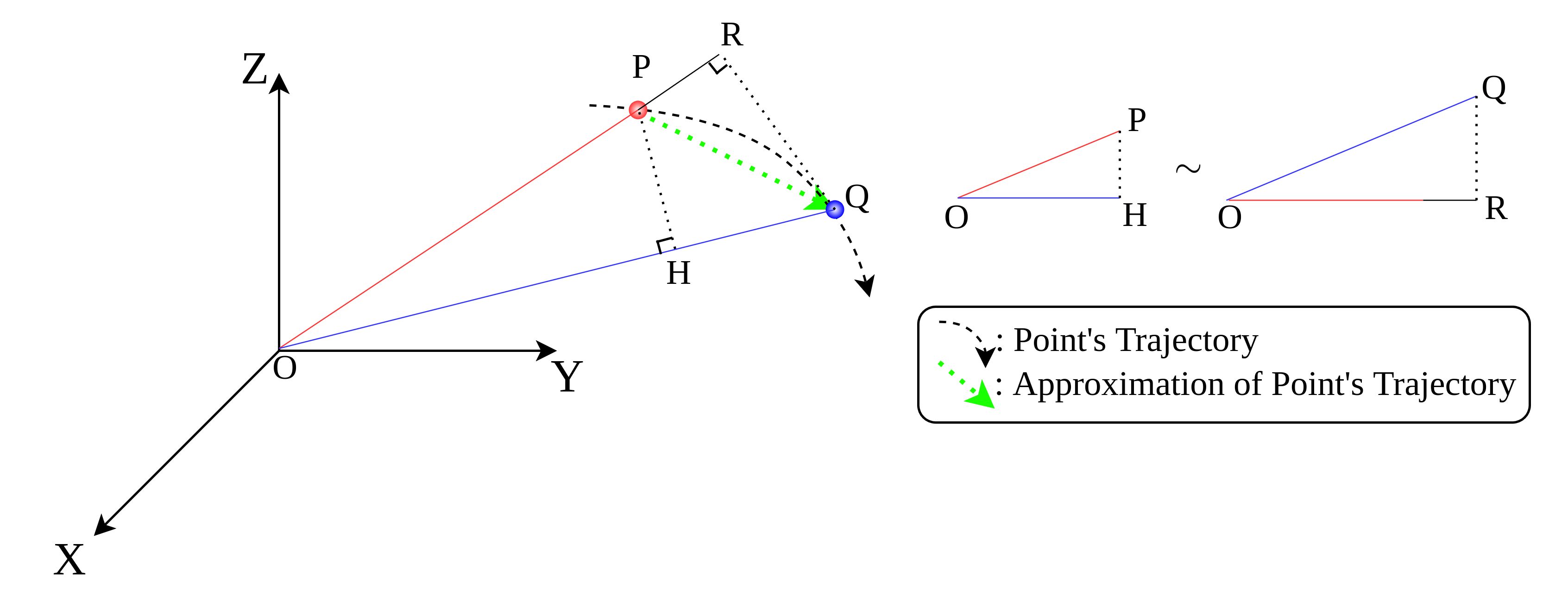}
	\caption{Two corresponding points across consecutive scans. 
		The radial and tangential velocity components form the basis of our Doppler Correspondence derivation.}
	\label{fig: Doppler Geometry}
\end{figure}

Let us denote points \(p_i \in \mathcal{P}\) and \(q_j \in \mathcal{Q}\) are the true corresponding points, represented as:
\[
p_i = \begin{bmatrix} x_{p,i} & y_{p,i} & z_{p,i} \end{bmatrix}^\top \quad \text{and} \quad 
q_j = \begin{bmatrix} x_{q,j} & y_{q,j} & z_{q,j} \end{bmatrix}^\top.
\]
Here, \((x, y, z)\) are the spatial coordinates of the point. Besides, let \(v_{p,i}\) and \(v_{q,j}\) be the Doppler velocity of each point. Assuming that, during a short time interval \(\Delta T\) between two consecutive scans, the trajectory of the point can be approximated by the vector \(\overrightarrow{PQ}\), as shown in Fig.~\ref{fig: Doppler Geometry}. Decomposing \(\overrightarrow{PQ}\) into radial (\(r\)) and tangential (\(\theta\)) components yields:
\[
\overrightarrow{PQ} = \overrightarrow{PR} + \overrightarrow{RQ},
\]
where \(\overrightarrow{PR}\) and \(\overrightarrow{RQ}\) represent the radial and tangential components, respectively. Expressing this in terms of velocity:
\[
\overrightarrow{V} \Delta T = \overrightarrow{V}_{p,r} \Delta T + \overrightarrow{V}_{p,\theta} \Delta T,
\]
where \(\overrightarrow{V}\) is the average velocity over \(\overrightarrow{PQ}\), and \(\overrightarrow{V}_{p,r}\) and \(\overrightarrow{V}_{p,\theta}\) are the radial and tangential velocities at point \(p_i\), respectively. Similarly, for \(q_i\):
\[
\overrightarrow{PQ} = \overrightarrow{HQ} + \overrightarrow{PH},
\]
and:
\[
\overrightarrow{V} \Delta T = \overrightarrow{V}_{q,r} \Delta T + \overrightarrow{V}_{q,\theta} \Delta T.
\]
Using the similarity condition \( \triangle POH \sim \triangle QOR \) (i.e., side-angle-side similarity), the following proportional relationship holds:
\[
\frac{\vert \overrightarrow{OR} \vert}{\vert \overrightarrow{OQ} \vert} = \frac{\vert \overrightarrow{OH} \vert}{\vert \overrightarrow{OP} \vert}.
\]
Expanding this gives:
\begin{equation} \label{eq:init}
\frac{\vert \overrightarrow{OP} \vert + \vert \overrightarrow{PR} \vert}{\vert \overrightarrow{OQ} \vert} = \frac{\vert \overrightarrow{OQ} \vert - \vert \overrightarrow{HQ} \vert}{\vert \overrightarrow{OP} \vert}.
\end{equation}
Substituting \(\vert \overrightarrow{OP} \vert = r_{p,i}\), \(\vert \overrightarrow{OQ} \vert = r_{q,j}\), \(\vert \overrightarrow{PR} \vert = v_{p,i} \Delta T\), and \(\vert \overrightarrow{HQ} \vert = v_{q,j} \Delta T\), where \(r_{p,i}\) and \(r_{q,j}\) are the ranges of \(p_i\) and \(q_j\) respectively, and \(v_{p,i}\) and \(v_{q,j}\) are their respective Doppler velocities, into Eq.~\eqref{eq:init} provides:
\[
\frac{r_{p,i} + v_{p,i} \Delta T}{r_{q,j}} = \frac{r_{q,j} - v_{q,j} \Delta T}{r_{p,i}}.
\]
Rearranging and simplifying the above equation yields:  
\begin{equation} \label{eq:rearranged}
	r_{p,i}^2 + r_{p,i} v_{p,i} \Delta T = r_{q,j}^2 - r_{q,j} v_{q,j} \Delta T.
\end{equation}
This can be expressed as:  
\begin{equation} \label{eq:functions}
	f(p_i,v_{p,i}) = g(q_j,v_{q,j}).
\end{equation}
Note that the above relationship, termed as \emph{Doppler Correspondence}, is formulated such that the left-hand side, \( f(p_i,v_{p,i}) \), depends only on the information from point \( p_i \), while the right-hand side, \( g(q_j,v_{q,j}) \), depends only on the information from point \( q_j  \). Furthermore, since there are no terms involving the \(\Delta t\), this Doppler Correspondence is inherently invariant to translation of the sensor. Moreover, unlike traditional correspondence that relies only on spatial \((x, y, z)\) information, this method additionally utilizes Doppler information. This enables a more robust and reliable matching process in repetitive geometric environments, as it relies more on the flow of the point cloud~\cite{ding2022self} rather than the spatial information from the environments.

\subsection{Kinematic Derivation of Doppler Correspondence}
Alternative to the previous geometric derivation of Doppler Correspondence, this section utilizes manipulation of simple matrices and vectors. To establish Doppler Correspondence, we start with the kinematic relationship from ~\cite{ding2022self, zhou2023self}:
\begin{align}
	\frac{q_j - p_i}{\Delta T} \cdot \frac{p_i}{\|p_i\|_2} &\approx v_{p,i}, \textsc{\label{eq:radial_p}} \\
	\frac{q_j - p_i}{\Delta T} \cdot \frac{q_j}{\|q_j\|_2} &\approx v_{q,j}. \label{eq:radial_q}
\end{align}
Here, each left-hand side of Eqs.~\eqref{eq:radial_p} and~\eqref{eq:radial_q}, represents the radial velocity computed relative to \(p_i\) and \(q_j\), respectively, while each right-hand side corresponds to the Doppler velocity measured by the sensor. By clearing denominators, we obtain:
\begin{align}
	(q_j - p_i) \cdot p_i &= v_{p,i} \,\Delta T\, \|p_i\|_2, 
	\label{eq:dot_product_p} \\
	(q_j - p_i) \cdot q_j &= v_{q,j} \,\Delta T\, \|q_j\|_2. 
	\label{eq:dot_product_q}
\end{align}
Adding Eqs.~\eqref{eq:dot_product_p} and~\eqref{eq:dot_product_q} yields:
\[
(q_j - p_i) \cdot (q_j + p_i) \;=\; v_{p,i} \,\Delta T\, \|p_i\|_2 \;+\; v_{q,j} \,\Delta T\, \|q_j\|_2.
\]
Since \((q_j - p_i) \cdot (q_j + p_i) \;=\; \|q_j\|_2^2 \;-\; \|p_i\|_2^2\), \(\|p_i\|_2 = r_{p,i}\), and \(\|q_j\|_2 = r_{q,j}\), where \(r_{p,i}\) and \(r_{q,j}\) are the ranges of \(p_i\) and \(q_j\), respectively, substituting these into the above equation leads to:
\[
r_{q,j}^2 \;-\; r_{p,i}^2 
= 
v_{p,i}\,r_{p,i}\,\Delta T 
\;+\; 
v_{q,j}\,r_{q,j}\,\Delta T.
\]
Rearranging terms, we obtain the final form:
\[
r_{p,i}^2 + r_{p,i} v_{p,i} \Delta T = r_{q,j}^2 - r_{q,j} v_{q,j} \Delta T.
\]
which matches Eq.~\eqref{eq:rearranged}.

\subsection{Outlier Rejection}
\begin{figure}[t]
	\centering
	\begin{subfigure}[ht]{0.35\textwidth} 
		\centering
		\includegraphics[width=\linewidth, height=5cm, keepaspectratio]{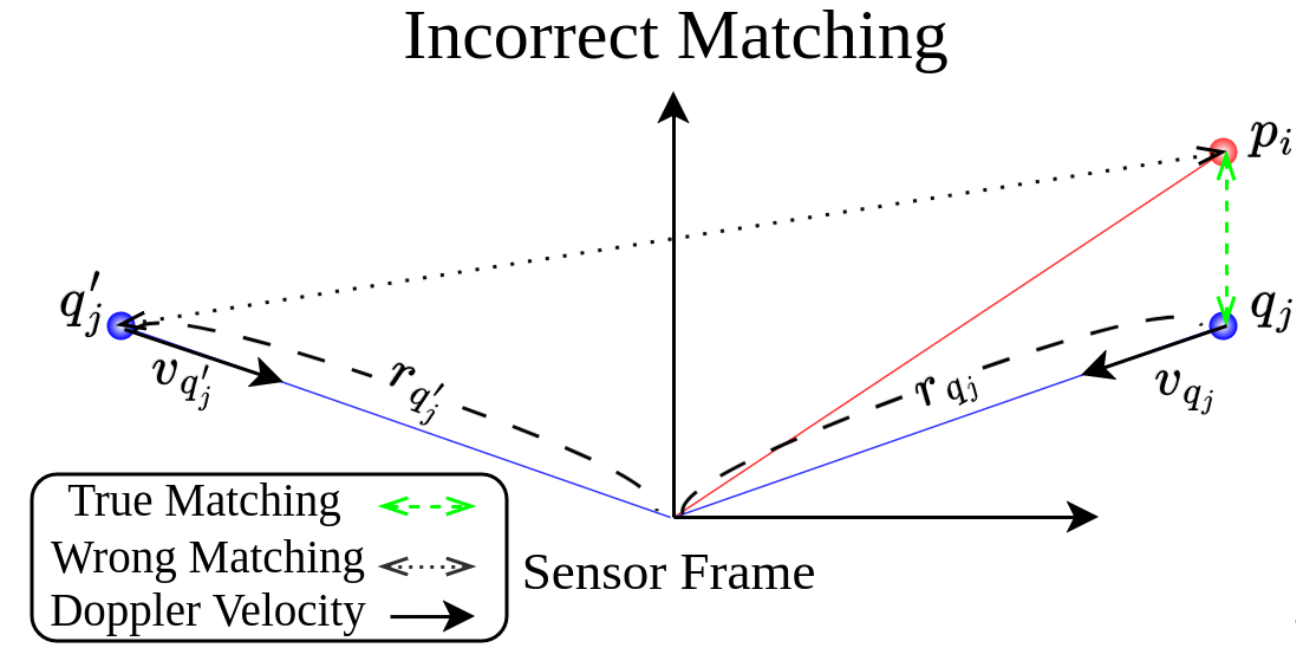}
		\caption{Incorrect matching}
	\end{subfigure}
	
	\begin{subfigure}[ht]{0.35\textwidth}
		\centering
		\includegraphics[width=\linewidth, height=5cm, keepaspectratio]{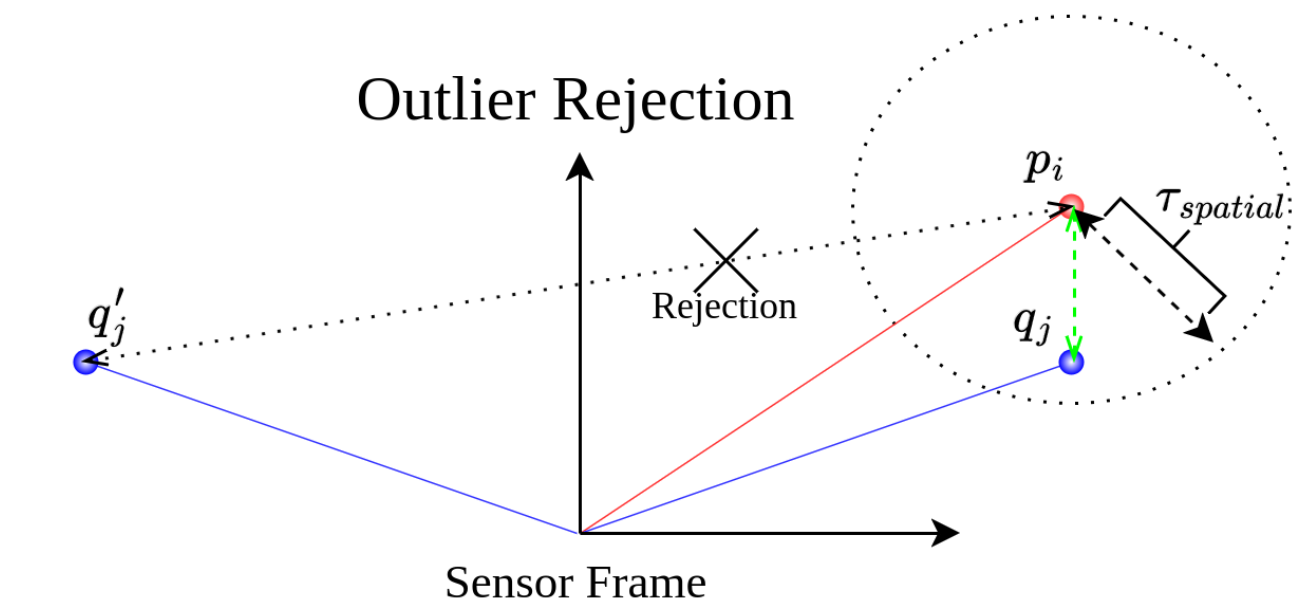}
		\caption{Outlier rejection}
	\end{subfigure}
	\caption{(a) illustrates multiple candidates due to symmetry in range and Doppler velocity, and (b) shows that the distance-based outlier rejection method filters out mismatched pairs.}
	\label{fig:wrong}
\end{figure}

A pair of points \( p_i \in \mathcal{P} \) and \( q_j \in \mathcal{Q} \) that satisfy Eq.~\eqref{eq:functions} can be considered as a correspondence. However, ambiguity could arise when multiple candidate points, such as \( q_j \) and its symmetric counterpart \( q'_j \in \mathcal{Q} \) with respect to the sensor frame, satisfy Doppler Correspondence for a single \( p_i \). In this case, both points yield \( f(p_i,v_{p,i}) = g(q_j,v_{q,j}) = g(q'_j,v_{q',j}) \), as shown in Fig.~\ref{fig:wrong}(a).

This occurs because \( g(q_j,v_{q,j}) \) is identical to \( g(q'_j,v_{q',j}) \), as the range and radial velocity of each point is the same (\( r_{q_j} = r_{q'_j} \text{ and } v_{q_j} = v_{q'_j}\)). Such ambiguity can result in incorrect correspondence estimation. To address this issue, it is assumed that correctly matched points \( p_i \) and \( q_j \) lie within similar spatial regions between scans. This assumption enables the use of a distance-based rejection mechanism to filter out mismatched candidates, as shown in Fig.~\ref{fig:wrong}(b). Similar to the maximum correspondence distance used in~\cite{segal2009generalized}, a threshold is applied to eliminate unlikely matches. Specifically, both the spatial distance \( d_{\text{spatial}} = \| p_i - q_j \|_2 \) and the Doppler distance \( d_{\text{doppler}} = |f(p_i,v_{p,i}) - g(q_j,v_{q,j})| \) should satisfy:

\[
d_{\text{spatial}} \leq \tau_{\text{spatial}} \quad \text{and} \quad d_{\text{doppler}} \leq \tau_{\text{doppler}}.
\]
Any pair of points that significantly deviates from these constraints is considered an outlier and is removed from the set of correspondences. By enforcing these thresholds, the impact of false correspondence estimation is mitigated.
\label{section3}
\label{subsec: derive}

\subsection{Non-iterative Scan Matching Algorithm}
In this section, a simple registration method leveraging Doppler Correspondence is introduced. This method is designed to evaluate the feasibility of the proposed correspondence for odometry. The framework is intentionally designed to remain lightweight to isolate and emphasize the impact of the proposed Doppler Correspondence. To focus solely on demonstrating the effectiveness of this correspondence, techniques such as robust kernels and RANSAC are deliberately omitted from the algorithm. Further details are provided in Algorithm~\ref{NICP}.

The process begins by receiving two consecutive scans, \(\mathcal{P}\) and \(\mathcal{Q}\), and calculating \(f(p_i,v_{p,i})\text{ and }g(q_j,v_{q,j})\). \texttt{FindClosest} is then performed to estimate correspondences. Unlike the conventional ICP explained in Algorithm~\ref{ICP}, which relies on 3D-based correspondence, our method operates on 1D-based correspondence. This could reduce computational complexity, even though \( f(p_i, v_{p,i}) \) and \( g(q_j, v_{q,j}) \) need to be calculated. While this approach may be slightly less accurate than 3D-based methods due to less information, it offers faster computations. Furthermore, unlike the closest point correspondence methods \cite{besl1992method, hexsel2022dicp, rusinkiewicz2001efficient, segal2009generalized}, which iteratively refine correspondences through multiple iterations, Algorithm~\ref{NICP} performs a direct correspondence estimation in a single step. This eliminates the iterative search process, significantly reducing the computational load, as illustrated in Fig.~\ref{fig: non iter}. Next section, the feasibility of Doppler Correspondence is validated by comparing the proposed method against ICP-based approaches.
\begin{algorithm}
	\caption{Non-Iterative Scan Matching Algorithm Based on Doppler Correspondence}
	\begin{algorithmic}[ht]
		\Require 
		$\mathcal{P}$ (source point cloud), 
		$\mathcal{Q}$ (target point cloud), 
		$\tau_{\text{spatial}}$ (spatial distance threshold), 
		$\tau_{\text{doppler}}$ (doppler distance threshold)
		\Ensure 
		$\hat{R}, \hat{t}$
		
		\State Compute $f(\mathcal{P}), g(\mathcal{Q})$
		\State $\mathcal{Q'} \gets \text{\texttt{FindClosest}}(f(\mathcal{P}), g(\mathcal{Q}))$
		\State \textbf{Masking Outliers:}
		\State \quad $\mathcal{M} \gets \|\mathcal{P} - \mathcal{Q'}\|_2 \leq \tau_{\text{spatial}} \text{ and } |f(\mathcal{P}) - g(\mathcal{Q'})| \leq \tau_{\text{doppler}}$
		\State \quad $\mathcal{C} \gets \mathcal{P}[\mathcal{M}], \mathcal{Q'}[\mathcal{M}]$
		\State \textbf{Predict Transformation:}
		\State \quad $\hat{R}, \hat{t} \gets \underset{R, t}{\arg\min} \sum_{(p, q) \in \mathcal{C}} \| q - (R p + t) \|_2$
		
		\Return $\hat{R}, \hat{t}$
	\end{algorithmic}
	\label{NICP}
\end{algorithm}
\begin{algorithm}
	\caption{Iterative Closest Point Algorithm}
	\begin{algorithmic}[ht]
		\Require 
		$\mathcal{P}$ (source point cloud), 
		$\mathcal{Q}$ (target point cloud), 
		$\tau_{\text{spatial}}$ (spatial distance threshold)
		\Ensure 
		$\hat{R}, \hat{t}$
		\While{not converged}
		\State $\mathcal{Q'} \gets \text{\texttt{FindClosest}}(\mathcal{P}, \mathcal{Q})$
		\State \textbf{Masking Outliers:}
		\State \quad $\mathcal{M} \gets \|\mathcal{P} - \mathcal{Q'}\|_2 \leq \tau_{\text{spatial}}$
		\State \quad $\mathcal{C} \gets \mathcal{P}[\mathcal{M}], \mathcal{Q'}[\mathcal{M}]$
		\State \textbf{Predict Transformation:}
		\State \quad $\delta{R}, \delta{t} \gets \underset{R, t}{\arg\min} \sum_{(p, q) \in \mathcal{C}} \| q - (R p + t) \|_2$
		\State \textbf{Update:}
		\State \quad $\mathcal{P} \gets \texttt{TransformSourcePoint}(\delta R, \delta t, \mathcal{P})$
		\State \quad $\hat{R},\hat{t}\gets \texttt{UpdateTransformation}(\delta{R},\delta{t},\hat{R},\hat{t})$
		\EndWhile
		
		\Return $\hat{R}, \hat{t}$
	\end{algorithmic}
	\label{ICP}
\end{algorithm}

\section{Experiment and Discussion}

\label{sec: exp}
\subsection{Dataset}
To evaluate the effectiveness of Doppler Correspondence for odometry, experiments are conducted on datasets with varying point cloud densities. The datasets include sparse 4D radar, semi-dense 4D radar, and dense 4D LiDAR point clouds. This diversity allows us to examine the performance of our approach across different sensing conditions.

\subsubsection{Sparse 4D Radar}
The \textit{Campus} dataset~\cite{li20234d} is utilized to evaluate sparse 4D radar point clouds. The data were collected using a ZF FRGen21 4D radar sensor mounted on the front bumper of a vehicle. Each 4D radar scan generates a sparse and noisy point cloud, containing approximately 400 to 1,400 points every 60 ms. The \textit{Campus} dataset encompasses outdoor environments featuring trees, vehicles, and pedestrians. These diverse settings challenge methods to operate effectively under low-density radar point cloud conditions.

\subsubsection{Semi-Dense 4D Radar}
The \textit{Loop 1} dataset from the NTU4DRadLM dataset~\cite{zhang2023ntu4dradlm} is used to evaluate semi-dense 4D radar. The data were collected using a car platform with an Oculii Eagle 4D radar, traversing the NTU campus main roads, resulting in a 6.95 km trajectory. The car moved at an average speed of 25--30 km/h, and the 4D radar scans exhibit semi-dense point clouds with approximately 1,000 to 2,000 points per 80ms. The environment features a mix of structured and semi-structured regions.

\subsubsection{Dense 4D LiDAR}
The \textit{Straight Wall}, \textit{Curved Wall}, and \textit{Baker-Barry Tunnel} datasets from \cite{hexsel2022dicp} are utilized to evaluate dense 4D LiDAR. The \textit{Straight Wall} and \textit{Curved Wall} datasets, generated in the CARLA simulator, simulate featureless environments by placing large parallel walls along the trajectory. These datasets provide an effective platform for evaluating odometry performance, particularly in environments with limited geometric features and repetitive structures. The \textit{Baker-Barry Tunnel} dataset was collected using an Aeva Aeries I FMCW 4D LiDAR sensor. The point clouds from 4D LiDAR are significantly denser than 4D radar data, containing approximately 40,000 to 120,000 points per 100ms, depending on the sequence.

\subsection{Experiment Setup}
The proposed method is evaluated by comparing its performance with established methods: point-to-point ICP and DICP~\cite{hexsel2022dicp}. DICP is a novel algorithm that integrates Doppler velocity into the \textbf{Optimization} in the ICP framework. It enhances scan matching by jointly optimizing both the closest point term and additional constraints derived from the Doppler velocity of each point. These methods are tested in two conditions: \textit{No Seed} and \textit{With Seed}. In the \textit{No Seed} category, no initial pose estimate is provided, whereas the \textit{With Seed} category uses the estimated pose from the previous scan pair under a constant-velocity motion model assumption~\cite{vizzo2023kiss}. 

For point-to-point ICP, the parameters are set to the following values: maximum iteration = 100, convergence threshold = $10^{-5}$, downsampling factor = 2, and $\tau_{\text{spatial}}$ = 3. The maximum iteration defines the maximum number of optimization iterations. The method is considered converged when the change in error falls below the convergence threshold. The point clouds are uniformly downsampled at intervals determined by the downsampling factor, reducing density and computational complexity.

DICP is evaluated using its original parameters from the open-source implementation. Each value is as follows: maximum iteration = 100, convergence threshold = $10^{-5}$, downsampling factor = 2, and $\tau_{\text{spatial}}$ = 0.3. 

The proposed method is evaluated with parameters set to these values: downsampling factor = 1, $\tau_{\text{spatial}}$ = 3, and $\tau_{\text{doppler}}$ = 5. It does not use maximum iteration or convergence threshold, as it performs optimization in a single step. All experiments are conducted on a system equipped with an Intel Core i7-14700KF CPU.

\begin{table*}[ht]
	\centering
	\scriptsize
	\setlength{\tabcolsep}{5pt}
	\setlength\dashlinedash{0.5pt}
	\setlength\dashlinegap{1.5pt}
	\setlength\arrayrulewidth{0.3pt}
	
	\caption{Each cell shows rotation and translation errors in degrees and meters, respectively for different segment lengths. The final column highlights the average computation time per frame (ms). For each segment, the bold values represent the best performance.}
	\vspace{-2.5mm}
	\begin{tabular}{@{}c c c c c c c c c c@{}}
		\toprule[1pt]
		& & & \multicolumn{7}{c}{\textbf{Evaluation Units (rotation [deg] / translation [m])}} \\ \cmidrule(lr){4-10}
		& \multicolumn{1}{c}{\raisebox{5pt}[0pt][0pt]{\normalsize {Dataset}}}
		& \multicolumn{1}{c}{\raisebox{5pt}[0pt][0pt]{\normalsize {Method}}}
		& \textbf{8m} & \textbf{16m} & \textbf{24m} & \textbf{32m} & \textbf{40m} & \textbf{48m} & \textbf{Time (ms)} \\ 
		\midrule[1pt]
		
		\multirow{6}{*}{%
			\rotatebox{90}{%
				\shortstack{\textbf{Sparse}\\\textbf{4D Radar}}
			}
		}
		& \multirow{3}{*}{\emph{Campus1}} 
		& (With Seed)point-to-point 
		& 3.776 / \textbf{0.937} & 4.543 / \textbf{1.708} & 5.223 / \textbf{2.452} & 5.881 / \textbf{3.176} & 6.519 / \textbf{3.914} & 7.144 / \textbf{4.654} & 5.4\\
		&  
		& (No Seed)point-to-point   
		& 3.666 / 2.293 & 4.411 / 4.413 & 5.028 / 6.436 & 5.666 / 8.368 & 6.305 / 10.245 & 6.895 / 12.087 & 6.0\\
		&  
		& Ours                      
		& \textbf{3.594} / 1.215 & \textbf{4.149} / 2.189 & \textbf{4.468} / 3.048 & \textbf{4.811} / 3.828 & \textbf{5.158} / 4.557 & \textbf{5.422} / 5.285 & \textbf{0.17}\\
		\cdashline{2-10}
		
		& \multirow{3}{*}{\emph{Campus2}} 
		& (With Seed)point-to-point 
		& 2.690 / 1.329 & 3.640 / 2.578 & 4.533 / 3.800 & 5.410 / 5.019 & 6.294 / 6.203 & 7.156 / 7.378 & 5.6\\
		&  
		& (No Seed)point-to-point   
		& 2.526 / 2.516 & 3.314 / 4.918 & 4.068 / 7.323 & 4.798 / 9.720 & 5.534 / 12.128 & 6.236 / 14.526 & 5.9\\
		&  
		& Ours                      
		& \textbf{2.379} / \textbf{0.769} & \textbf{2.957} / \textbf{1.271} & \textbf{3.505} / \textbf{1.707} & \textbf{4.090} / \textbf{2.078} & \textbf{4.712} / \textbf{2.401} & \textbf{5.278} / \textbf{2.680} & \textbf{0.16}\\
		\cdashline{1-10}
		
		\multirow{12}{*}{%
			\rotatebox{90}{%
				\shortstack{\textbf{Semi-Dense}\\\textbf{4D Radar}}
			}
		}
		& \multirow{3}{*}{\emph{Loop1\_0}} 
		& (With Seed)DICP          
		& 0.821 / \textbf{0.627} & 1.201 / 1.110 & 1.434 / 1.606 & 1.600 / 2.084 & 1.771 / 2.559 & 1.949 / 3.024 & 4.7\\
		&                 
		& (No Seed)DICP            
		& \textbf{0.813} / 0.631 & \textbf{1.193} / 1.112 & \textbf{1.419} / 1.605 & \textbf{1.590} / 2.084 & \textbf{1.763} / 2.566 & \textbf{1.935} / 3.033 & 4.8\\
		&                 
		& Ours                     
		& 1.172 / 0.646 & 1.750 / \textbf{1.071} & 2.242 / \textbf{1.500} & 2.649 / \textbf{1.906} & 3.044 / \textbf{2.295} & 3.421 / \textbf{2.734} & \textbf{1.2}\\
		\cdashline{2-10}
		
		& \multirow{3}{*}{\emph{Loop1\_2}} 
		& (With Seed)DICP          
		& 0.625 / \textbf{0.620} & \textbf{0.928} / \textbf{1.115} & \textbf{1.196} / \textbf{1.633} & \textbf{1.442} / 2.156 & \textbf{1.670} / 2.682 & \textbf{1.859} / 3.213 & 4.0\\
		&                 
		& (No Seed)DICP            
		& 0.625 / 0.623 & 0.932 / 1.124 & 1.207 / 1.647 & 1.461 / 2.177 & 1.698 / 2.709 & 1.897 / 3.247 & 4.4\\
		&                 
		& Ours                     
		& \textbf{0.617} / 0.650 & 0.942 / 1.148 & 1.216 / 1.651 & 1.472 / \textbf{2.145} & 1.707 / \textbf{2.631} & 1.913 / \textbf{3.110} & \textbf{0.4}\\
		\cdashline{2-10}
		
		& \multirow{3}{*}{\emph{Loop1\_3}} 
		& (With Seed)DICP          
		& \textbf{0.841} / 0.631 & 1.161 / \textbf{1.117} & \textbf{1.413} / \textbf{1.617} & \textbf{1.619} / 2.106 & \textbf{1.808} / 2.597 & \textbf{1.997} / 3.095 & 4.5\\
		&                 
		& (No Seed)DICP            
		& 0.843 / \textbf{0.630} & \textbf{1.160} / 1.119 & 1.415 / 1.623 & 1.622 / 2.118 & 1.812 / 2.615 & 2.003 / 3.118 & 4.6\\
		&                 
		& Ours                     
		& 1.396 / 0.704 & 2.138 / 1.204 & 2.650 / 1.687 & 3.044 / \textbf{2.106} & 3.399 / \textbf{2.502} & 3.768 / \textbf{2.882} & \textbf{1.2}\\
		\cdashline{2-10}
		
		& \multirow{3}{*}{\emph{Loop1\_4}} 
		& (With Seed)DICP          
		& 1.318 / \textbf{0.660} & 1.904 / 1.181 & 2.280 / 1.651 & 2.629 / 2.092 & 3.172 / 2.525 & 3.767 / 2.925 & 5.3\\
		&                 
		& (No Seed)DICP            
		& \textbf{1.253} / 0.663 & \textbf{1.807} / \textbf{1.170} & \textbf{2.160} / \textbf{1.633} & \textbf{2.452} / \textbf{2.065} & \textbf{2.963} / \textbf{2.477} & \textbf{3.540} / \textbf{2.827} & 5.3\\
		&                 
		& Ours                     
		& 3.123 / 0.880 & 5.398 / 1.760 & 7.349 / 2.700 & 9.169 / 3.782 & 11.949 / 5.227 & 15.604 / 6.883 & \textbf{2.2}\\
		\cdashline{1-10}
		
		\multirow{12}{*}{%
			\rotatebox{90}{%
				\shortstack{\textbf{Dense}\\\textbf{4D LiDAR}}
			}
		}
		& \multirow{3}{*}{\emph{Straight Wall}} 
		& (With Seed)DICP   
		& \textbf{0.050} / \textbf{0.028} & \textbf{0.111} / \textbf{0.059} & \textbf{0.128} / \textbf{0.073} & \textbf{0.154} / \textbf{0.086} & \textbf{0.258} / \textbf{0.103} & \textbf{0.303} / \textbf{0.112} & 29.1\\
		&                      
		& (No Seed)DICP     
		& 0.050 / 0.028 & 0.111 / 0.059 & 0.128 / 0.073 & 0.154 / 0.086 & 0.258 / 0.103 & 0.304 / 0.112 & 29.1\\
		&                      
		& Ours              
		& 0.486 / 0.068 & 0.896 / 0.173 & 1.420 / 0.291 & 1.882 / 0.525 & 2.213 / 0.789 & 2.627 / 1.091 & \textbf{7.8}\\
		\cdashline{2-10}
		
		& \multirow{3}{*}{\emph{Curved Wall}} 
		& (With Seed)DICP   
		& 0.378 / 0.092 & 0.761 / 0.137 & 1.167 / 0.133 & 1.565 / 0.120 & 1.974 / 0.223 & 2.399 / 0.404 & 29.1\\
		&                    
		& (No Seed)DICP     
		& 0.371 / \textbf{0.090} & 0.745 / \textbf{0.135} & 1.143 / \textbf{0.131} & 1.532 / \textbf{0.119} & 1.932 / \textbf{0.220} & 2.347 / \textbf{0.397} & 29.2\\
		&                    
		& Ours              
		& \textbf{0.327} / 0.183 & \textbf{0.601} / 0.368 & \textbf{0.856} / 0.552 & \textbf{1.069} / 0.724 & \textbf{1.261} / 0.888 & \textbf{1.442} / 1.050 & \textbf{8.6}\\
		\cdashline{2-10}
		
		& \multirow{3}{*}{\emph{Baker-Barry Tunnel (Empty)}}
		& (With Seed)DICP   
		& 1.028 / \textbf{0.289} & 1.652 / \textbf{0.490} & 2.515 / \textbf{0.744} & 3.313 / \textbf{0.963} & 3.964 / \textbf{1.140} & 4.794 / \textbf{1.342} & 20.8\\
		&                
		& (No Seed)DICP     
		& \textbf{1.022} / 0.290 & \textbf{1.634} / 0.492 & \textbf{2.487} / 0.746 & \textbf{3.269} / 0.965 & \textbf{3.914} / 1.141 & \textbf{4.730} / 1.341 & 26.1\\
		&                
		& Ours              
		& 1.804 / 0.614 & 2.599 / 0.945 & 3.325 / 1.301 & 3.767 / 1.581 & 4.189 / 1.919 & 4.931 / 2.367 & \textbf{5.8}\\
		\cdashline{2-10}
		
		& \multirow{3}{*}{\emph{Baker-Barry Tunnel (Vehicles)}}
		& (With Seed)DICP   
		& \textbf{0.778} / \textbf{0.179} & \textbf{1.435} / \textbf{0.339} & \textbf{2.110} / \textbf{0.497} & \textbf{2.525} / \textbf{0.621} & \textbf{3.027} / \textbf{0.732} & \textbf{3.479} / 0.833 & 20.1\\
		&                
		& (No Seed)DICP     
		& 0.780 / 0.181 & 1.449 / 0.341 & 2.121 / 0.499 & 2.542 / 0.624 & 3.049 / 0.733 & 3.506 / \textbf{0.831} & 24.1\\
		&                
		& Ours              
		& 2.216 / 0.776 & 3.659 / 1.417 & 4.904 / 2.170 & 6.099 / 2.633 & 7.167 / 3.138 & 8.121 / 3.646 & \textbf{3.6}\\
		\bottomrule[1pt]
	\end{tabular}
	\label{table:results}
\end{table*}
\begin{figure*}[ht]
	\vspace{-3mm}
	\centering
	\begin{subfigure}[t]{0.185\textwidth}
		\centering
		\includegraphics[width=\linewidth]{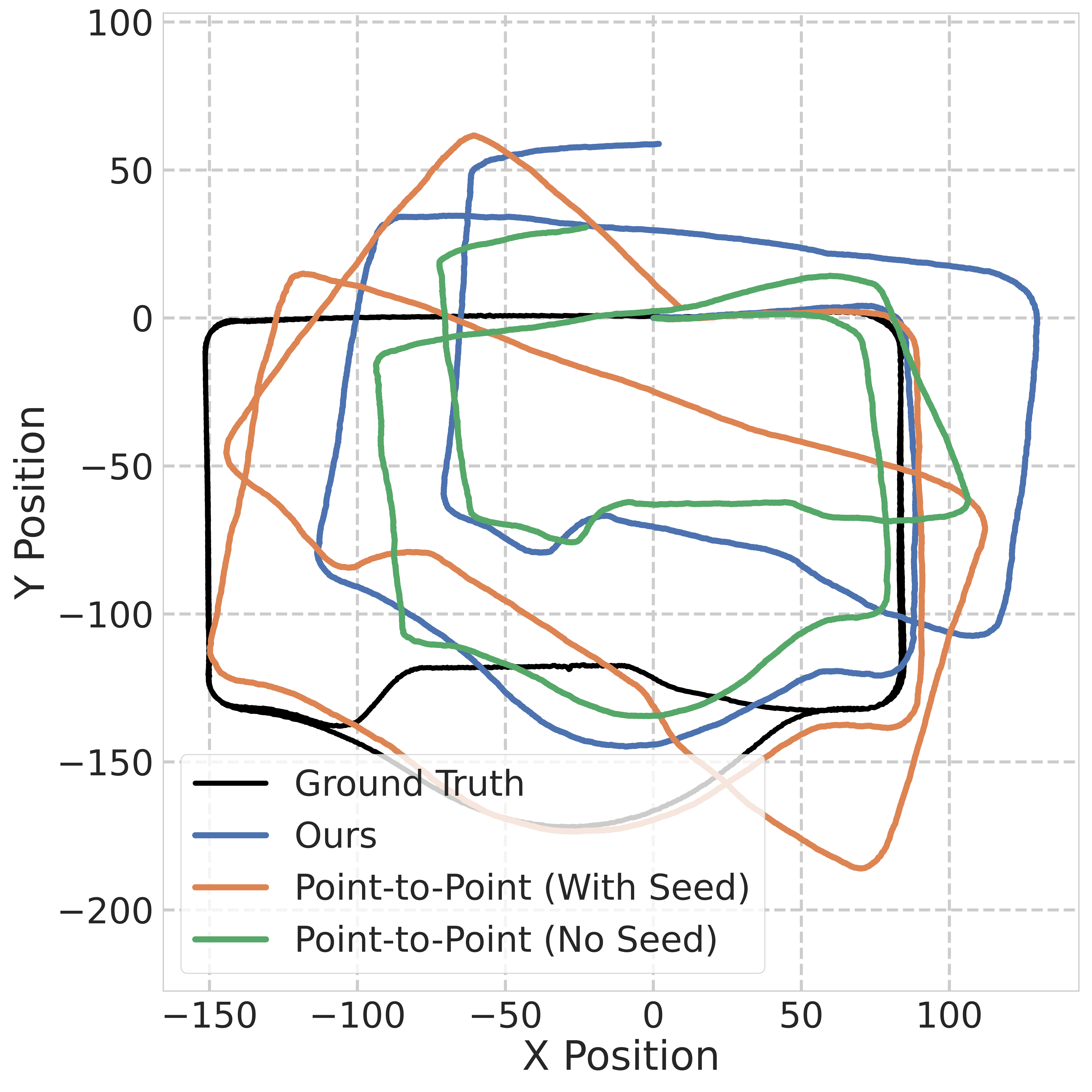}
		\caption{\textit{Campus1}}
	\end{subfigure}
	\hfill
	\begin{subfigure}[t]{0.185\textwidth}
		\centering
		\includegraphics[width=\linewidth]{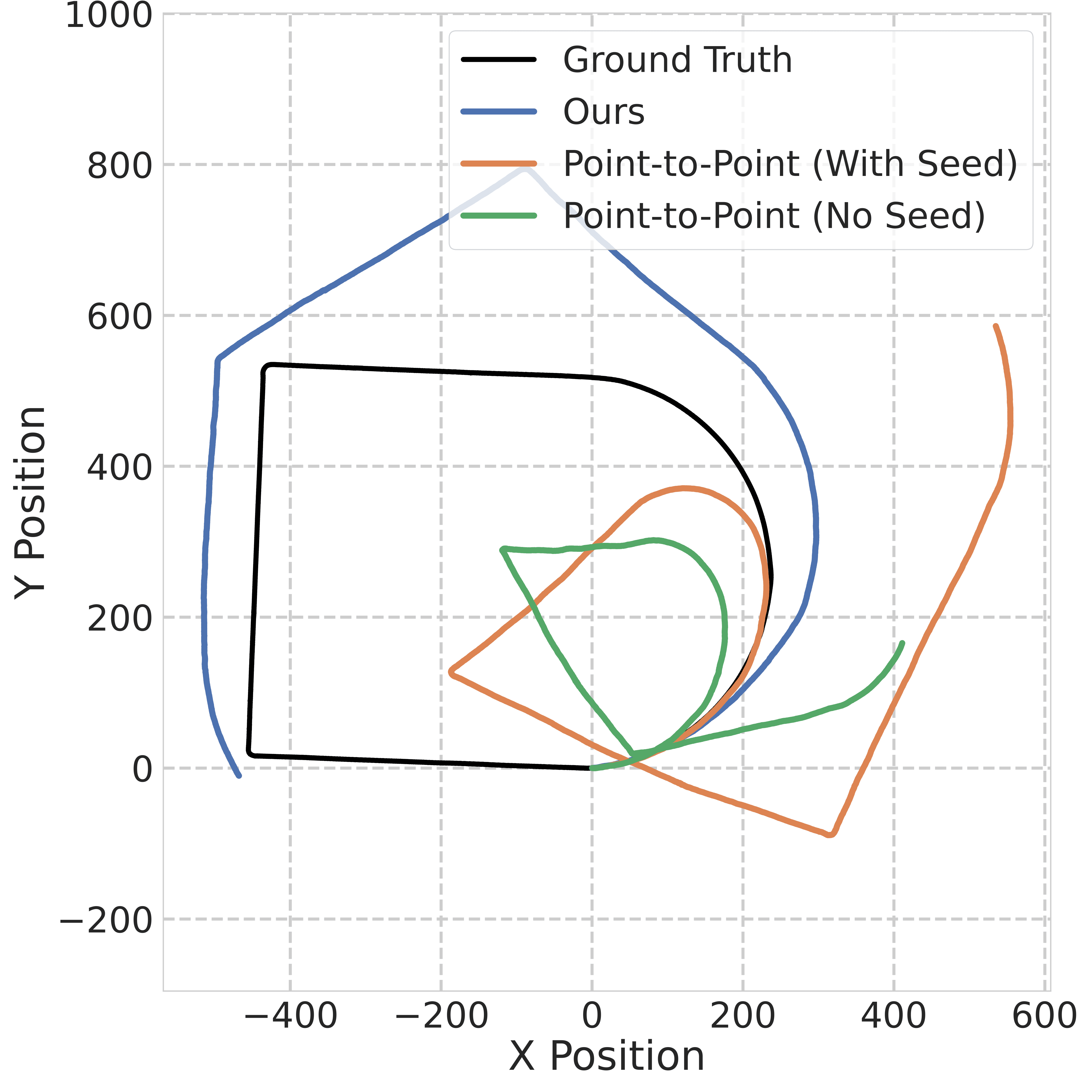}
		\caption{\textit{Campus2}}
	\end{subfigure}
	\hfill
	\begin{subfigure}[t]{0.185\textwidth}
		\centering
		\includegraphics[width=\linewidth]{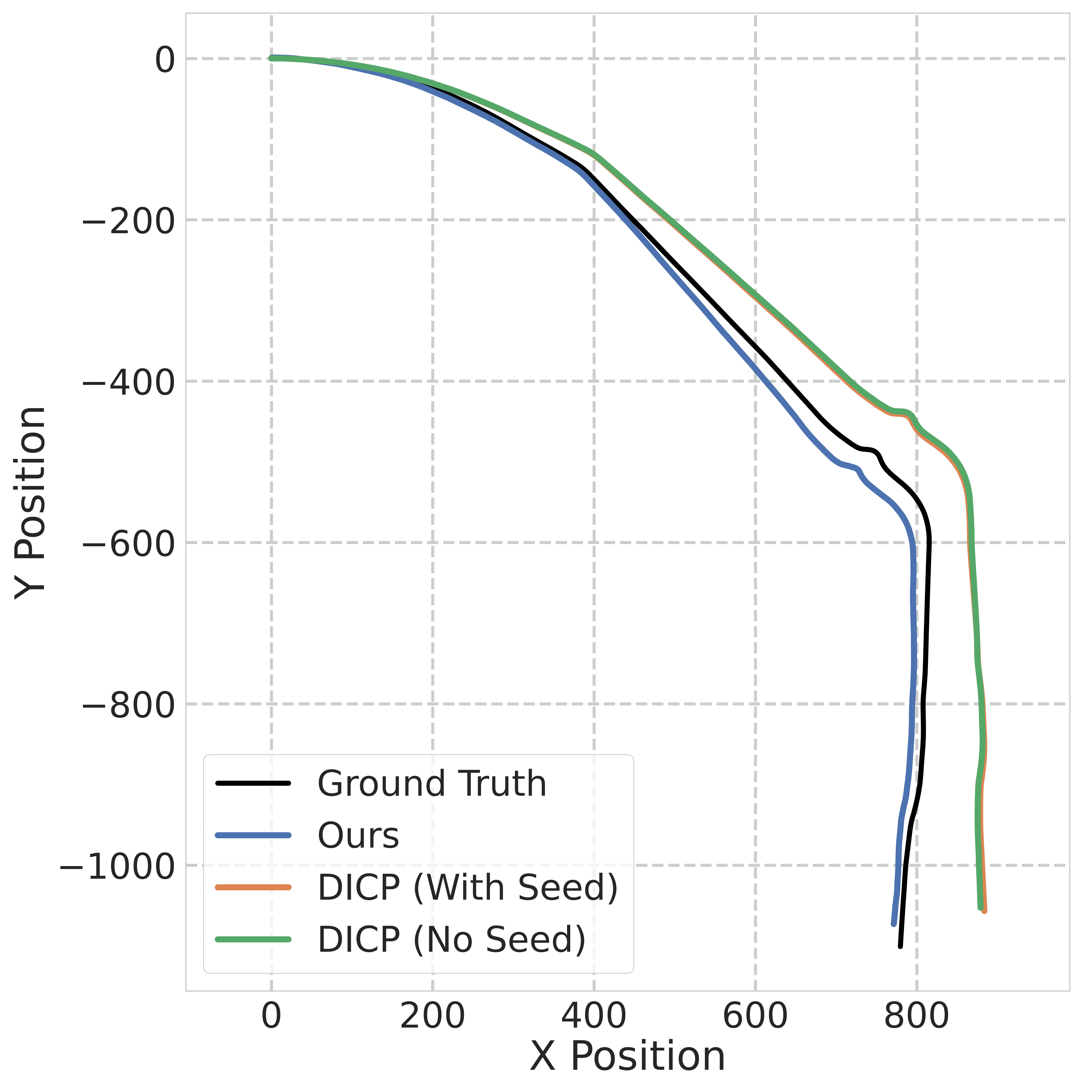}
		\caption{\textit{Loop1\_0}}
	\end{subfigure}
	\hfill
	\begin{subfigure}[t]{0.185\textwidth}
		\centering
		\includegraphics[width=\linewidth]{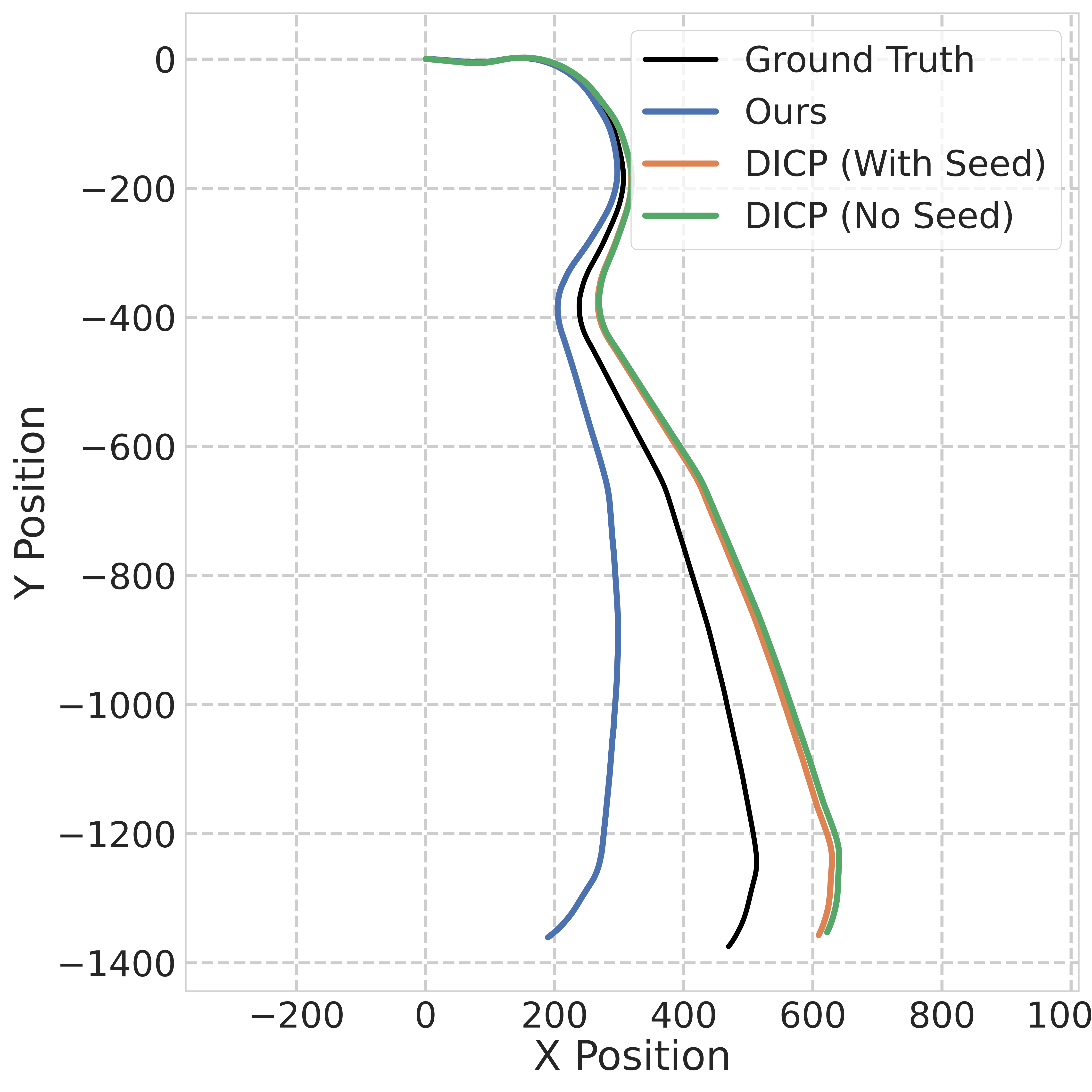}
		\caption{\textit{Loop1\_2}}
	\end{subfigure}
	\hfill
	\begin{subfigure}[t]{0.185\textwidth}
		\centering
		\includegraphics[width=\linewidth]{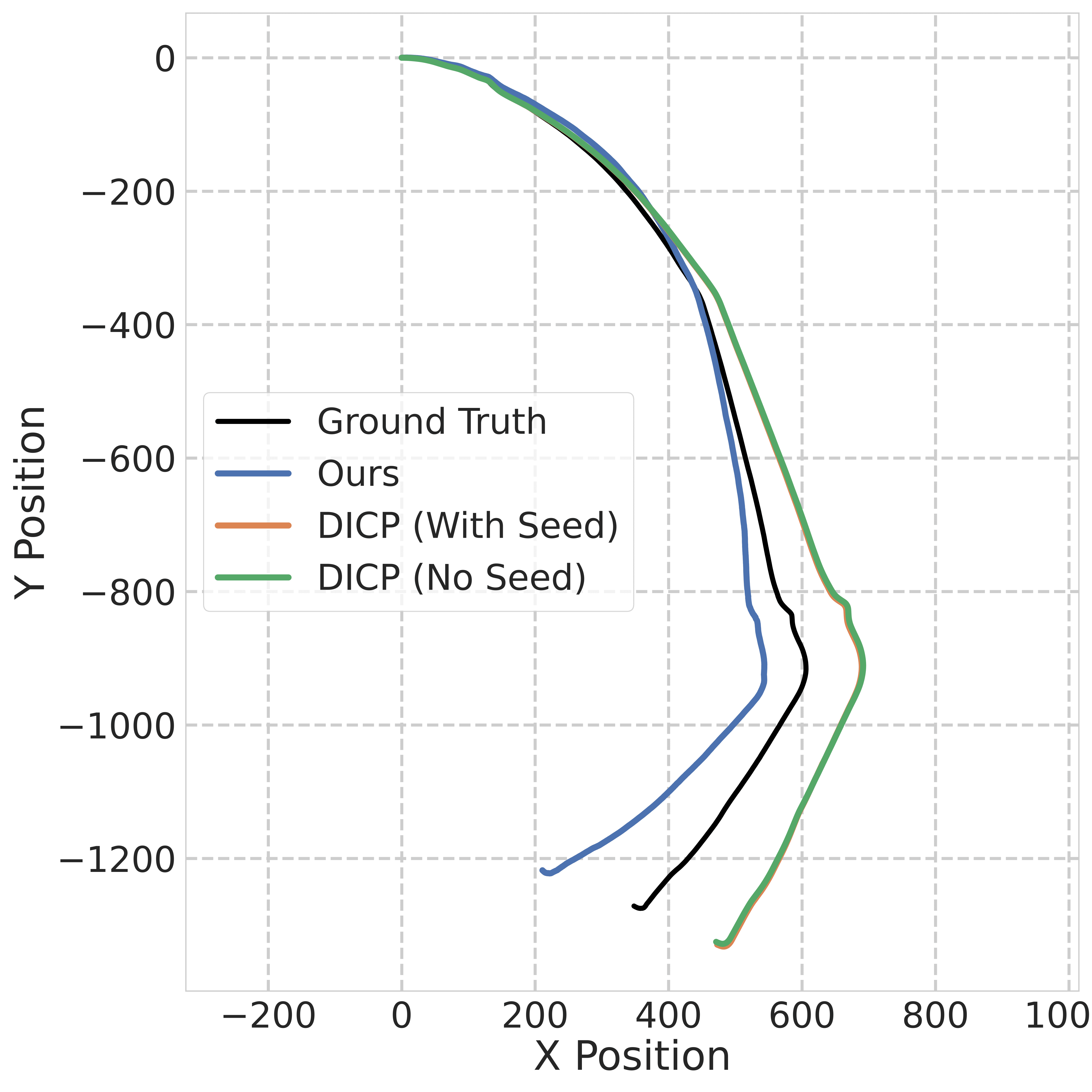}
		\caption{\textit{Loop1\_3}}
	\end{subfigure}
	\begin{subfigure}[t]{0.185\textwidth}
		\centering
		\includegraphics[width=\linewidth]{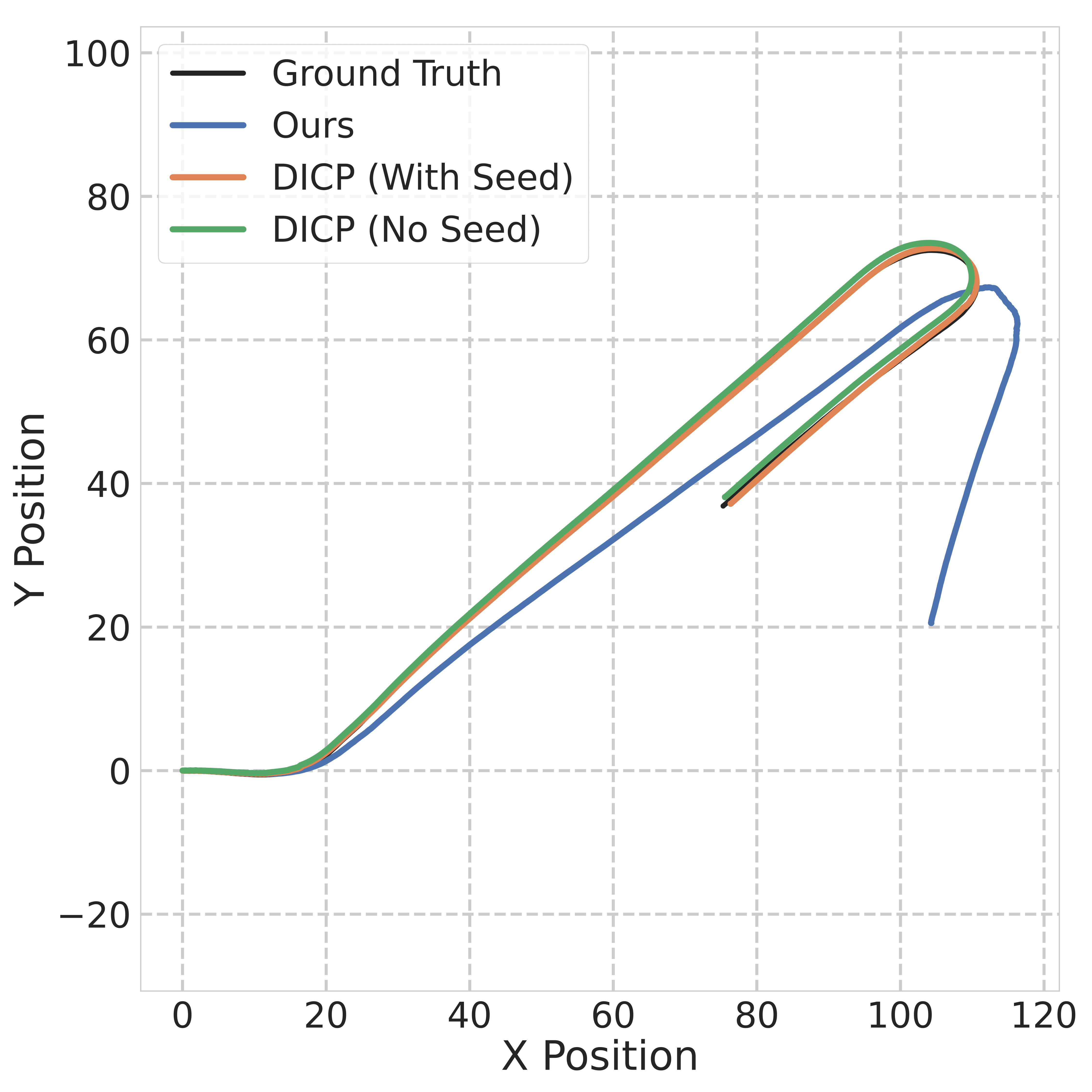}
		\caption{\textit{Loop1\_4}}
	\end{subfigure}
	\hfill
	\begin{subfigure}[t]{0.185\textwidth}
		\centering
		\includegraphics[width=\linewidth]{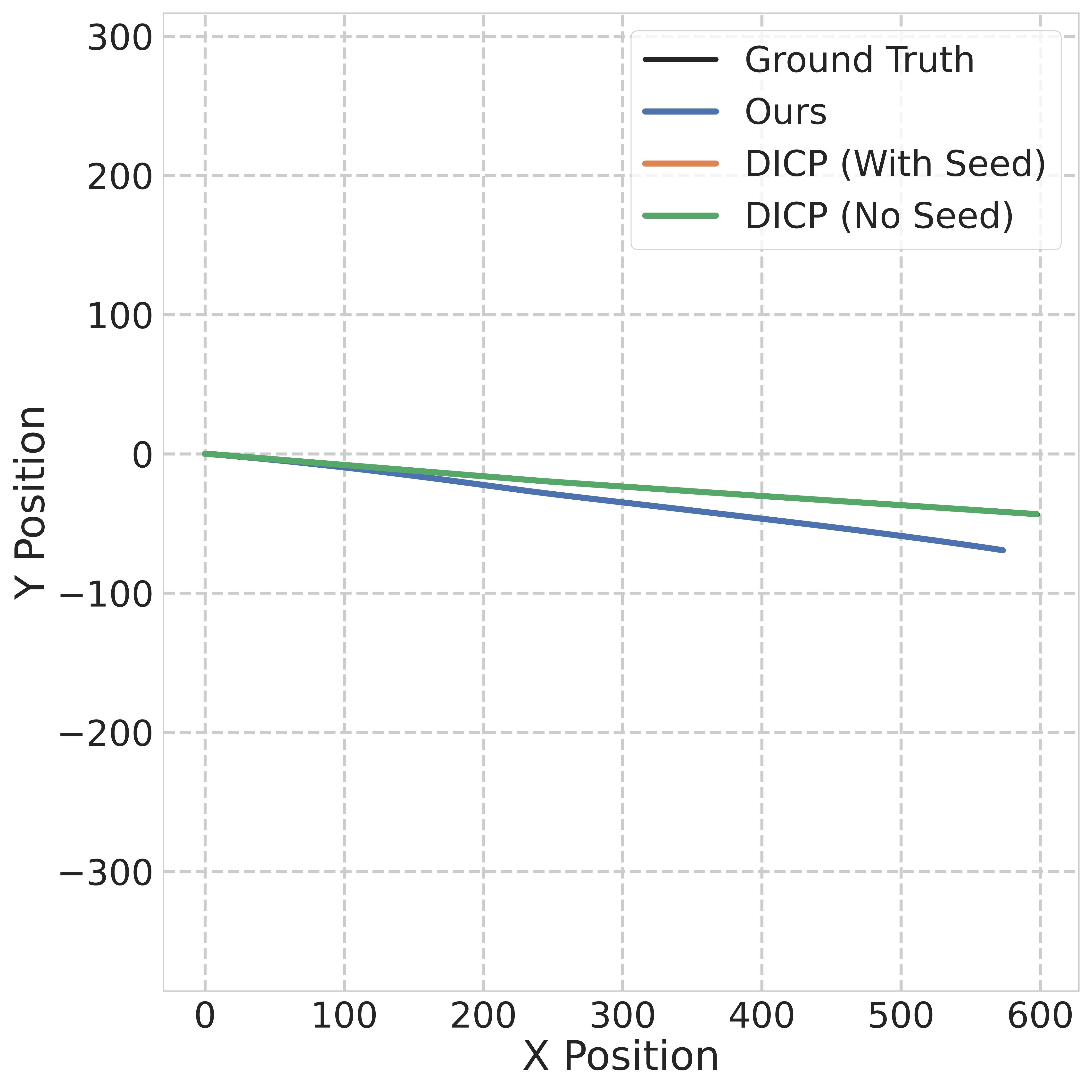}
		\caption{\textit{Straight Wall}}
	\end{subfigure}
	\hfill
	\begin{subfigure}[t]{0.185\textwidth}
		\centering
		\includegraphics[width=\linewidth]{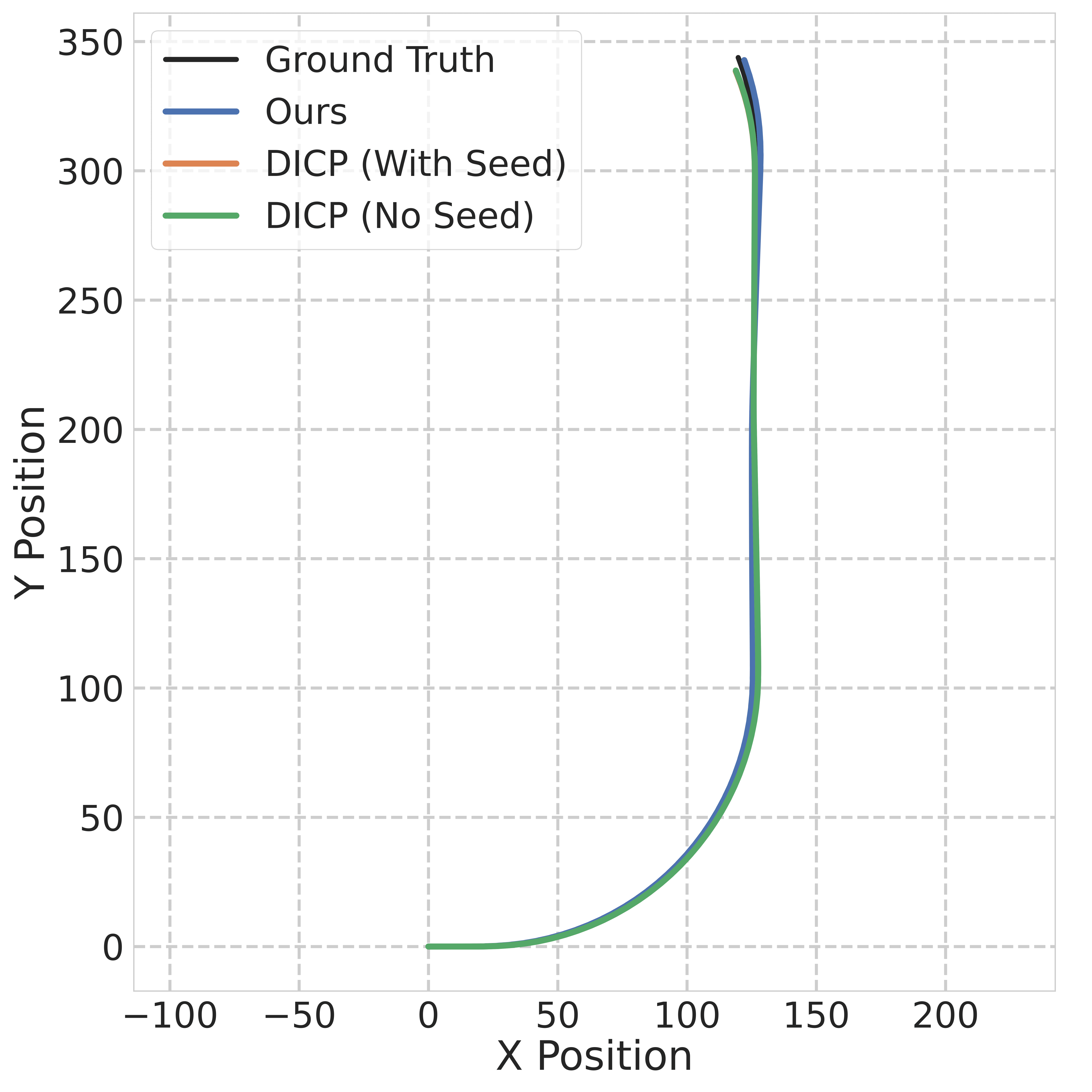}
		\caption{\textit{Curved Wall}}
	\end{subfigure}
	\hfill
	\begin{subfigure}[t]{0.185\textwidth}
		\centering
		\includegraphics[width=\linewidth]{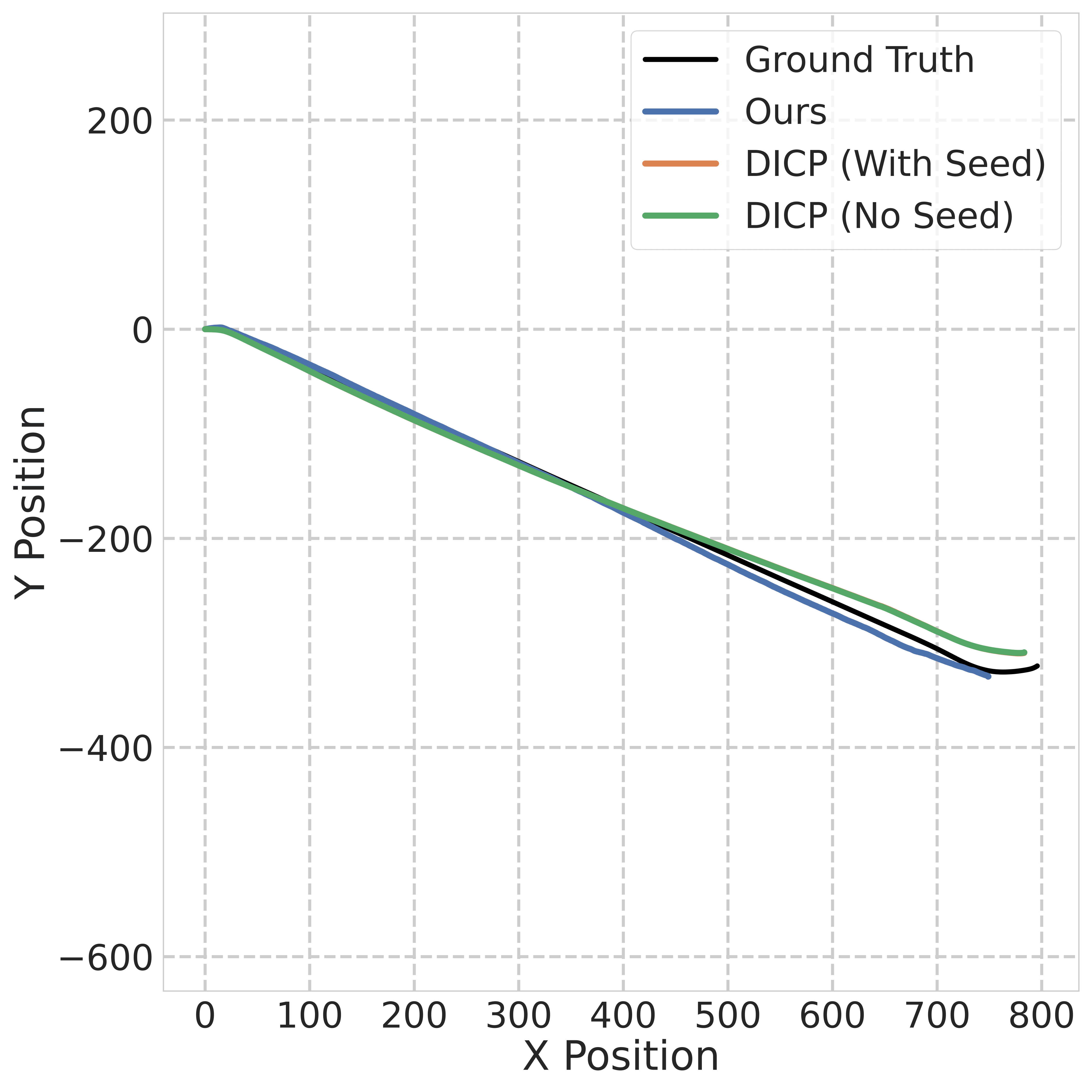}
		\caption{\parbox{0.8\linewidth}{\centering \textit{Baker-Barry\\ Tunnel (Empty)}}}
	\end{subfigure}
	\hfill
	\begin{subfigure}[t]{0.185\textwidth}
		\centering
		\includegraphics[width=\linewidth]{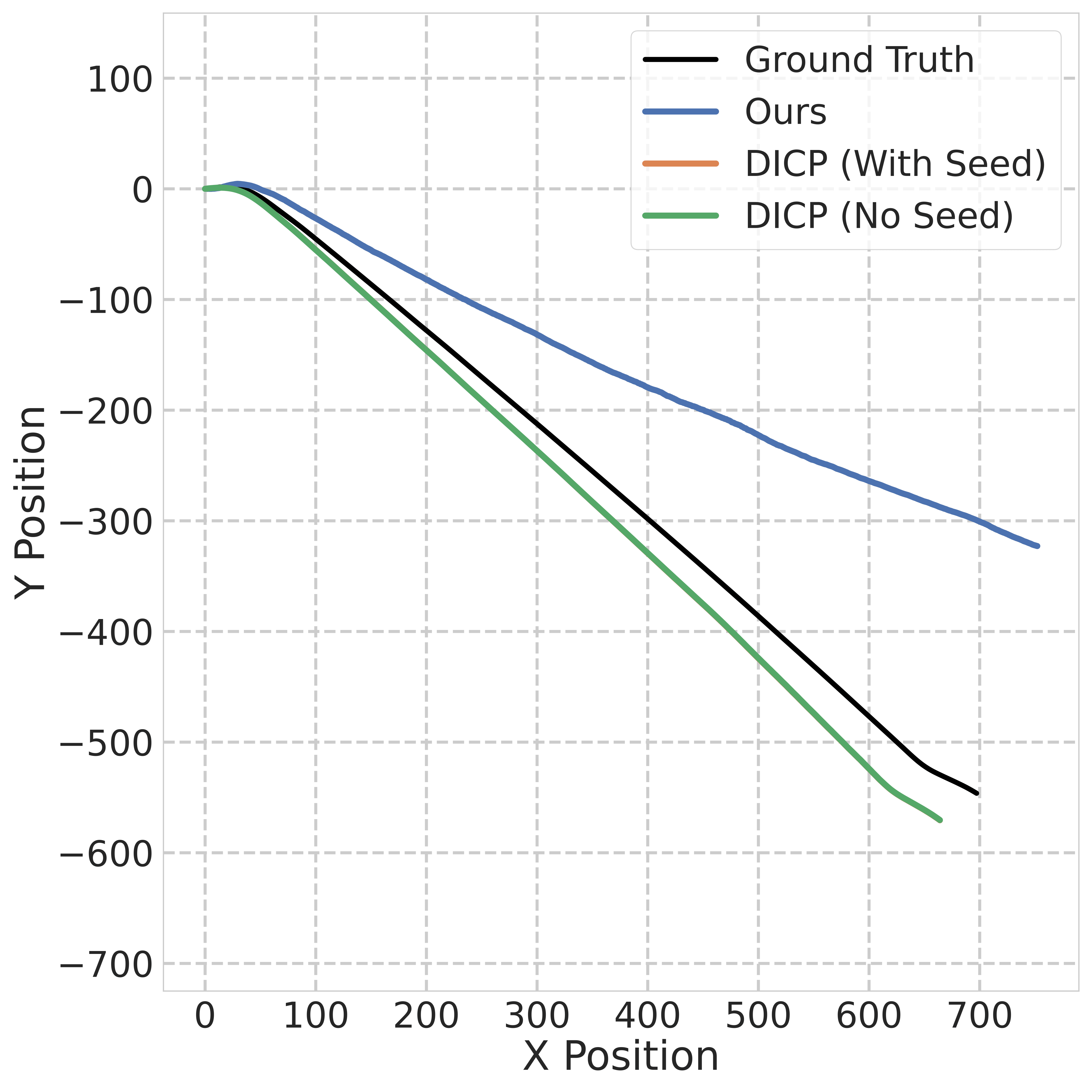}
		\caption{\parbox{0.8\linewidth}{\centering \textit{Baker-Barry\\ Tunnel (Vehicles)}}}
		
	\end{subfigure}
	
	\caption{Trajectory results on multiple sequences. 
		Each plot’s horizontal axis (x-axis) and vertical axis (y-axis) are in meters.}
	
	\label{fig:trajectory}
	\vspace{-5mm}
\end{figure*}

\subsection{Evaluation Metrics}
The performance is evaluated using the relative error (RE) from~\cite{zhang2018tutorial}. This metric standardizes trajectory estimation accuracy, allowing fair comparisons across different methods. The evaluation measures odometry accuracy through segment-wise error analysis. This approach provides detailed insights into the local and global performance of the trajectory estimation. By using shorter sub-trajectories, RE captures local consistency and short-term accuracy, while longer sub-trajectories emphasize global consistency and long-term accuracy. This dual capability makes RE a versatile and informative metric for trajectory analysis.

The computation time for transformation estimation, excluding point cloud pre-processing, is evaluated. A fair assessment of computational efficiency could be ensured by excluding computation time for point cloud pre-processing.

\begin{figure*}[h]
	\centering
	\begin{subfigure}[t]{0.32\textwidth}
		\includegraphics[width=\textwidth]{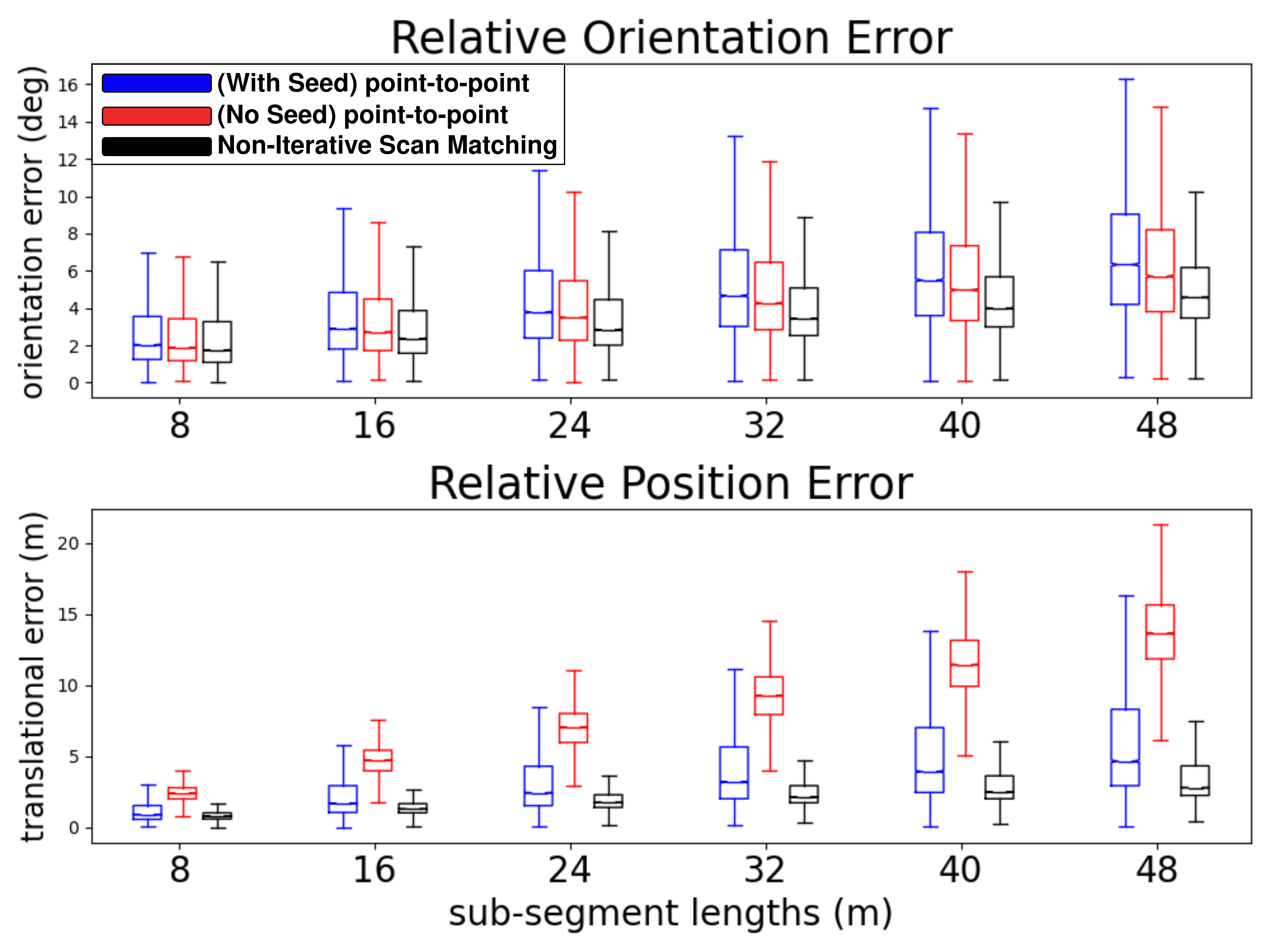}
		\caption{Sparse 4D Radar}
		\label{fig:error_comparsion_spare}
	\end{subfigure}
	\hfill
	\begin{subfigure}[t]{0.32\textwidth}
		\includegraphics[width=\textwidth]{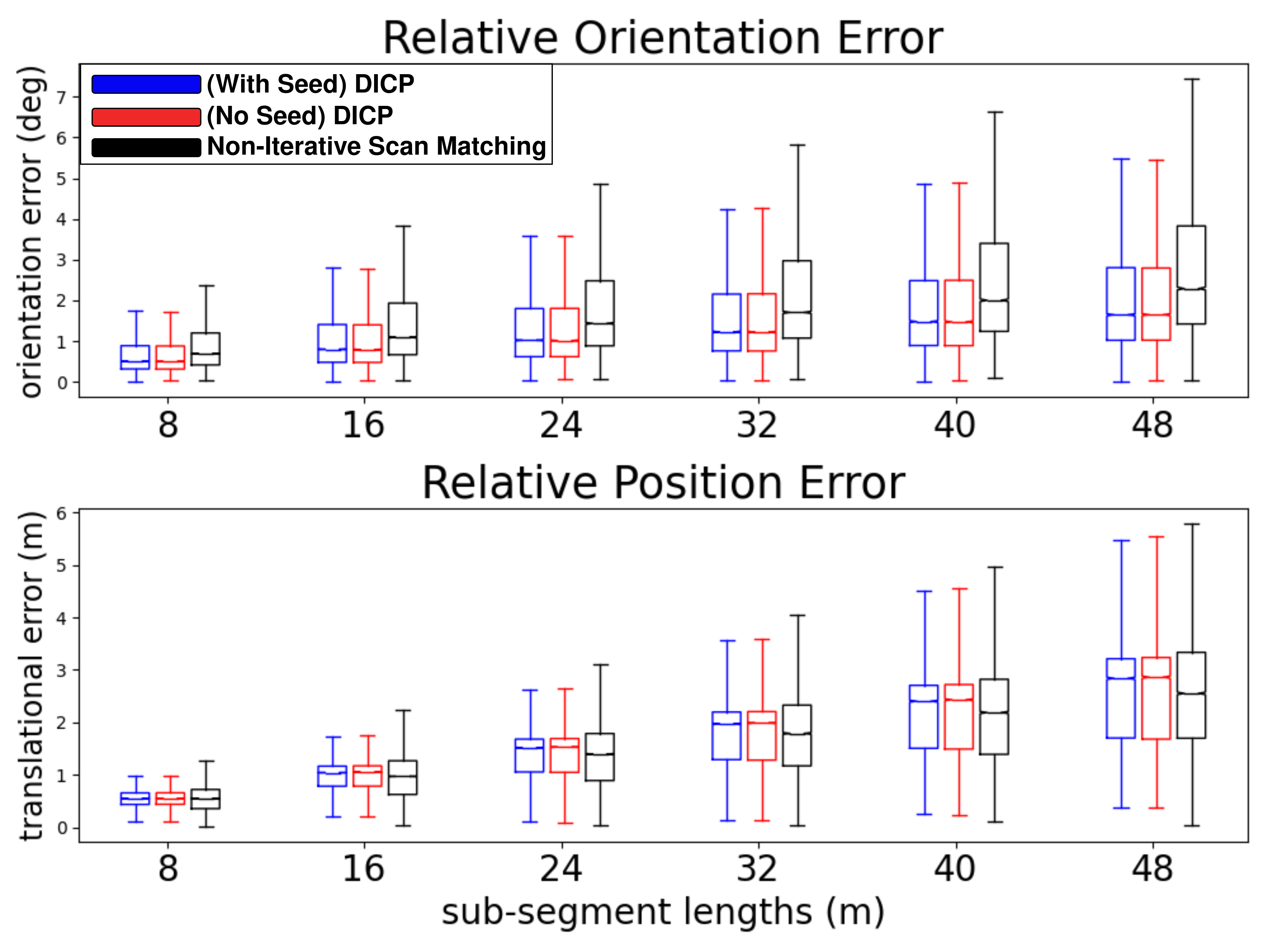}
		\caption{Semi-Dense 4D Radar}
	\end{subfigure}
	\hfill
	\begin{subfigure}[t]{0.32\textwidth}
		\includegraphics[width=\textwidth]{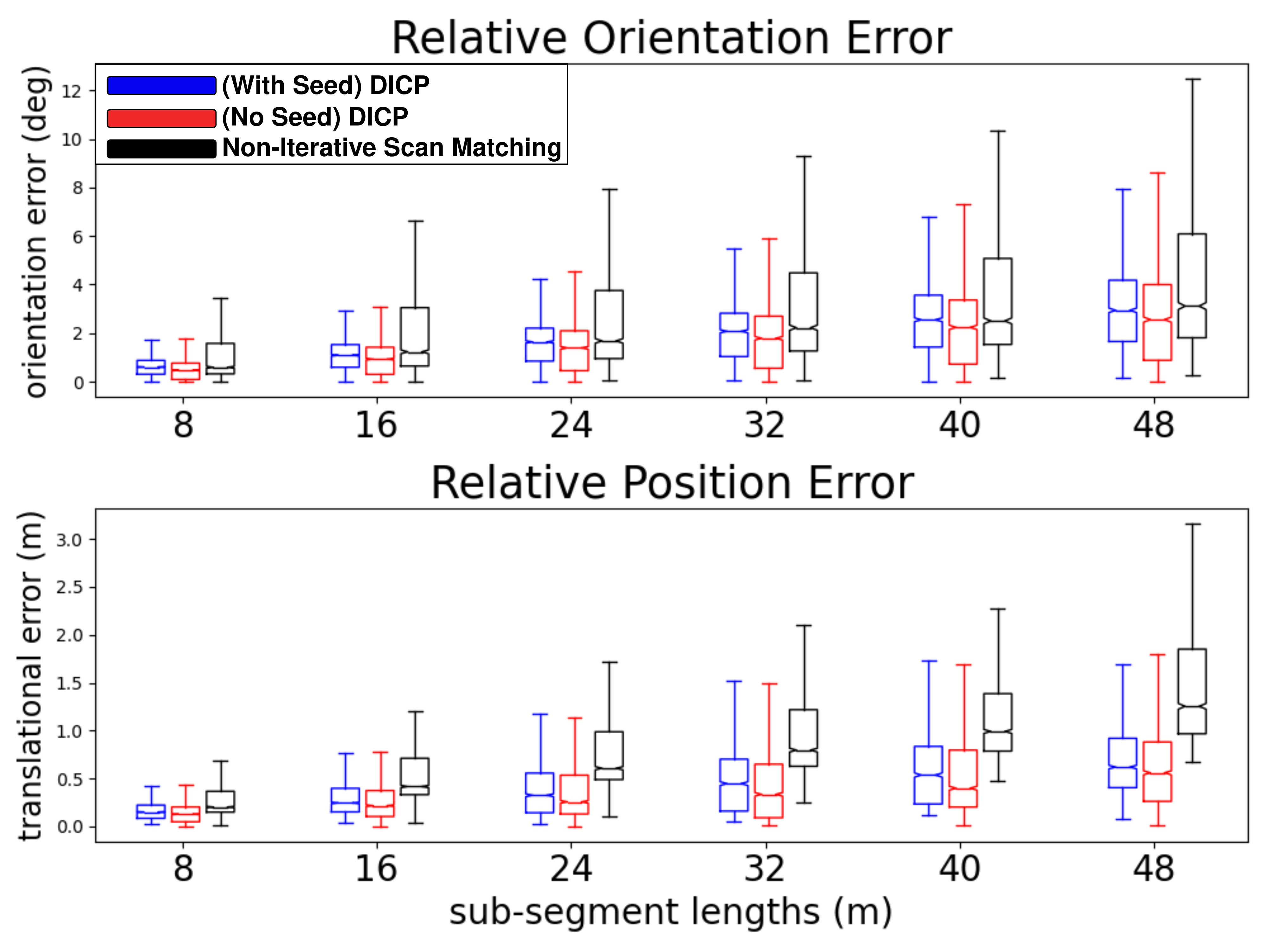}
		\caption{Dense 4D LiDAR}
	\end{subfigure}
	\caption{Error comparison of different methods, ICP, DICP, and Non-Iterative Scan Matching (Ours), across various point cloud densities. Each figure represents the averaged results for the corresponding dataset.}
	\label{fig: Error_Comparsion}
\end{figure*}
\begin{figure*}[h]
	\centering
	\begin{subfigure}[t]{0.3\textwidth}
		\centering
		\includegraphics[width=\linewidth]{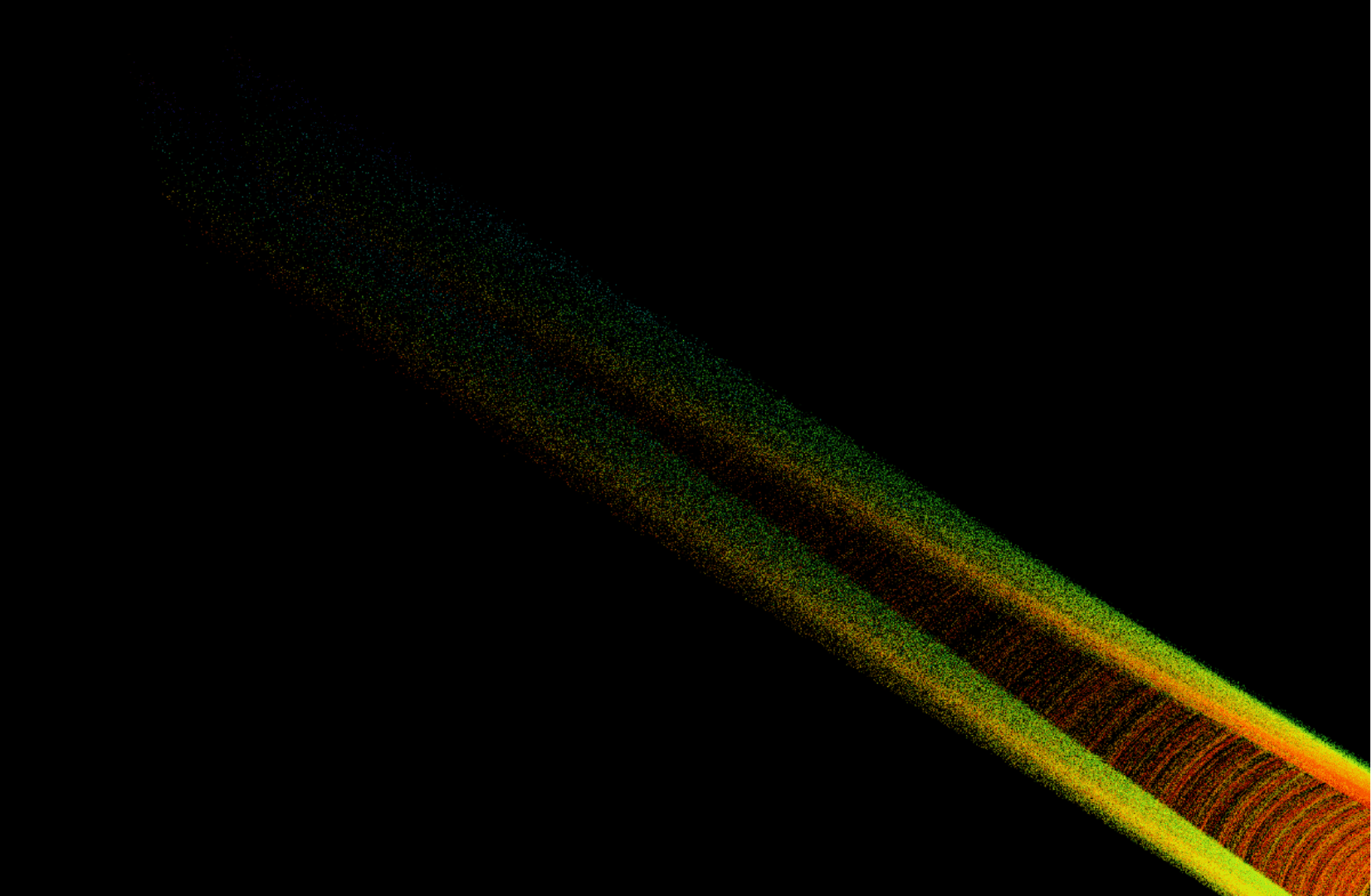}
		\caption{{ICP, \textit{Straight Wall}}}
	\end{subfigure}
	\hfill
	\begin{subfigure}[t]{0.3\textwidth}
		\centering
		\includegraphics[width=\linewidth]{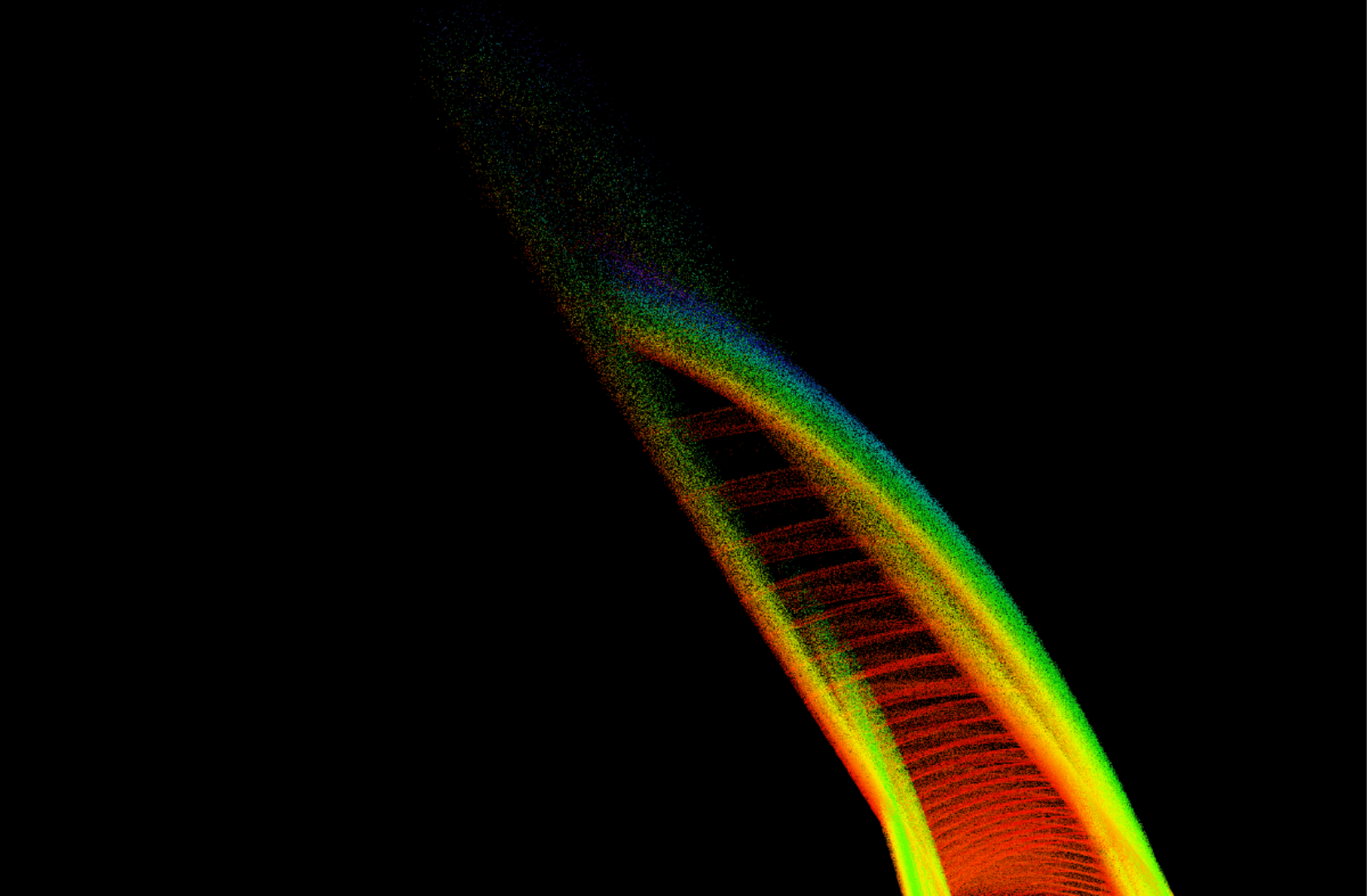}
		\caption{{ICP, \textit{Curved Wall}}}
	\end{subfigure}
	\hfill
	\begin{subfigure}[t]{0.3\textwidth}
		\centering
		\includegraphics[width=\linewidth]{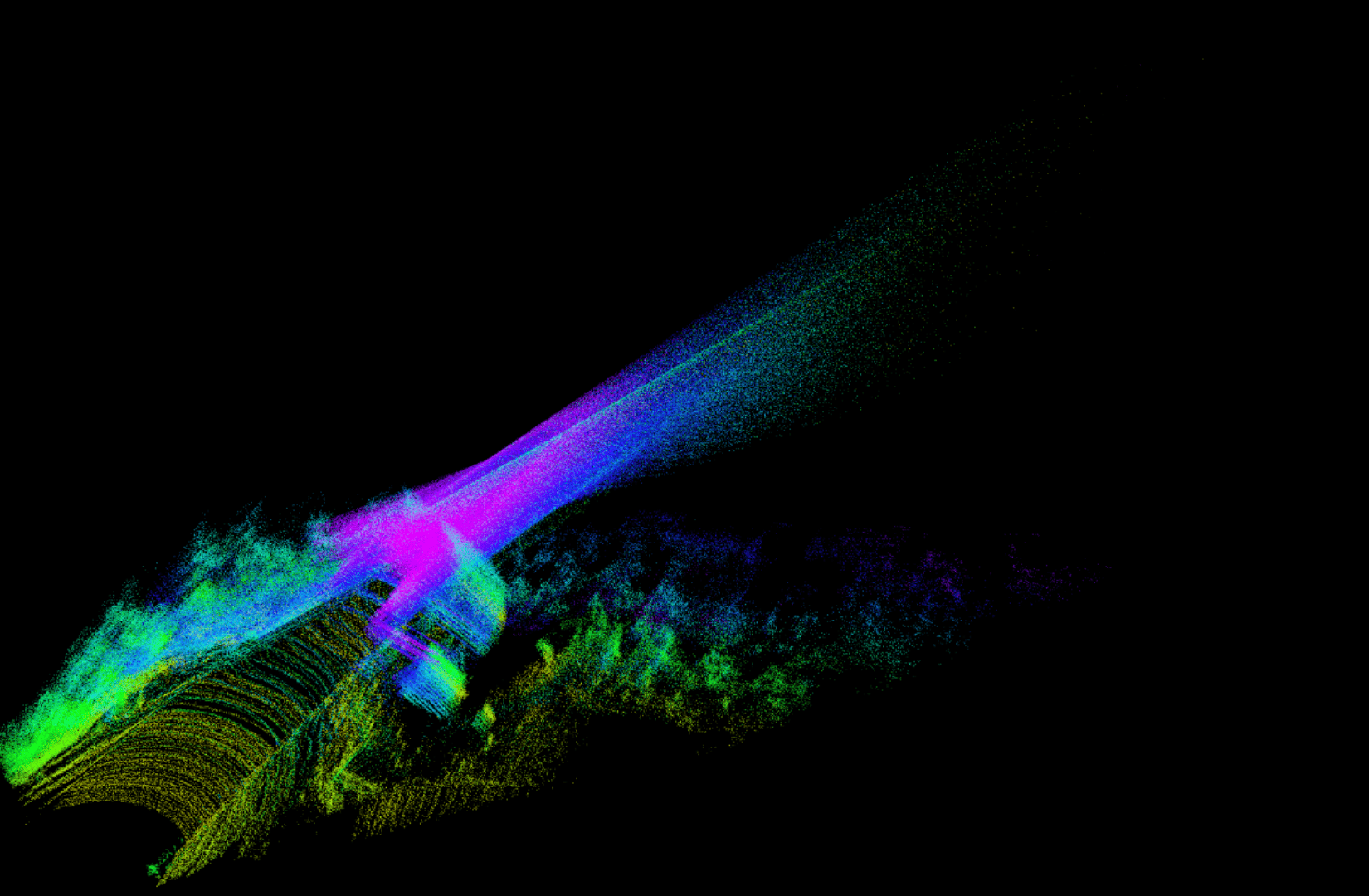}
		\caption{{ICP, \textit{Baker-Barry Tunnel (Empty)}}}
	\end{subfigure}
	\begin{subfigure}[t]{0.3\textwidth}
		\centering
		\includegraphics[width=\linewidth]{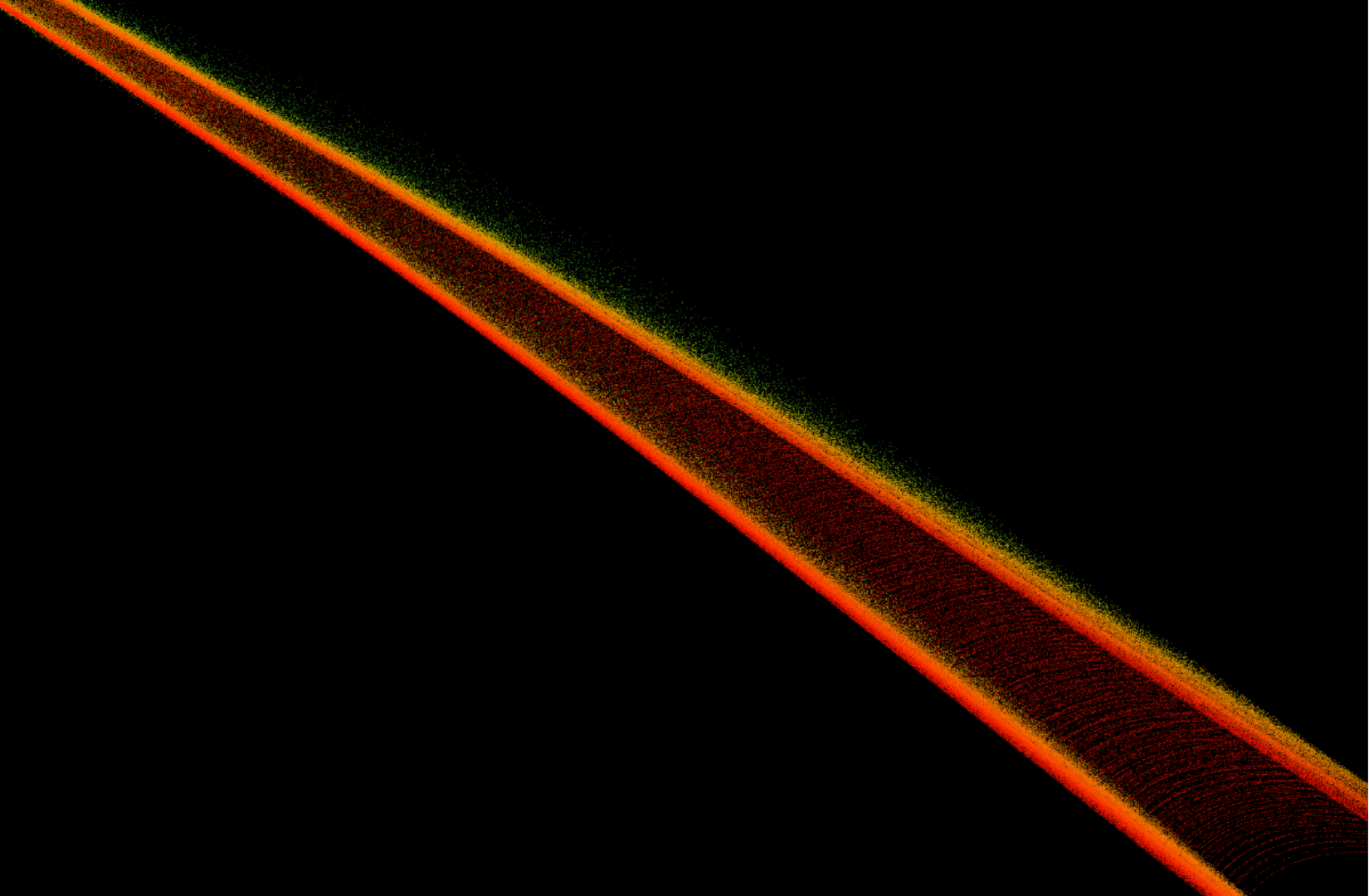}
		\caption{{Non-Iterative Scan Matching, \textit{Straight Wall}}}
	\end{subfigure}
	\hfill
	\begin{subfigure}[t]{0.3\textwidth}
		\centering
		\includegraphics[width=\linewidth]{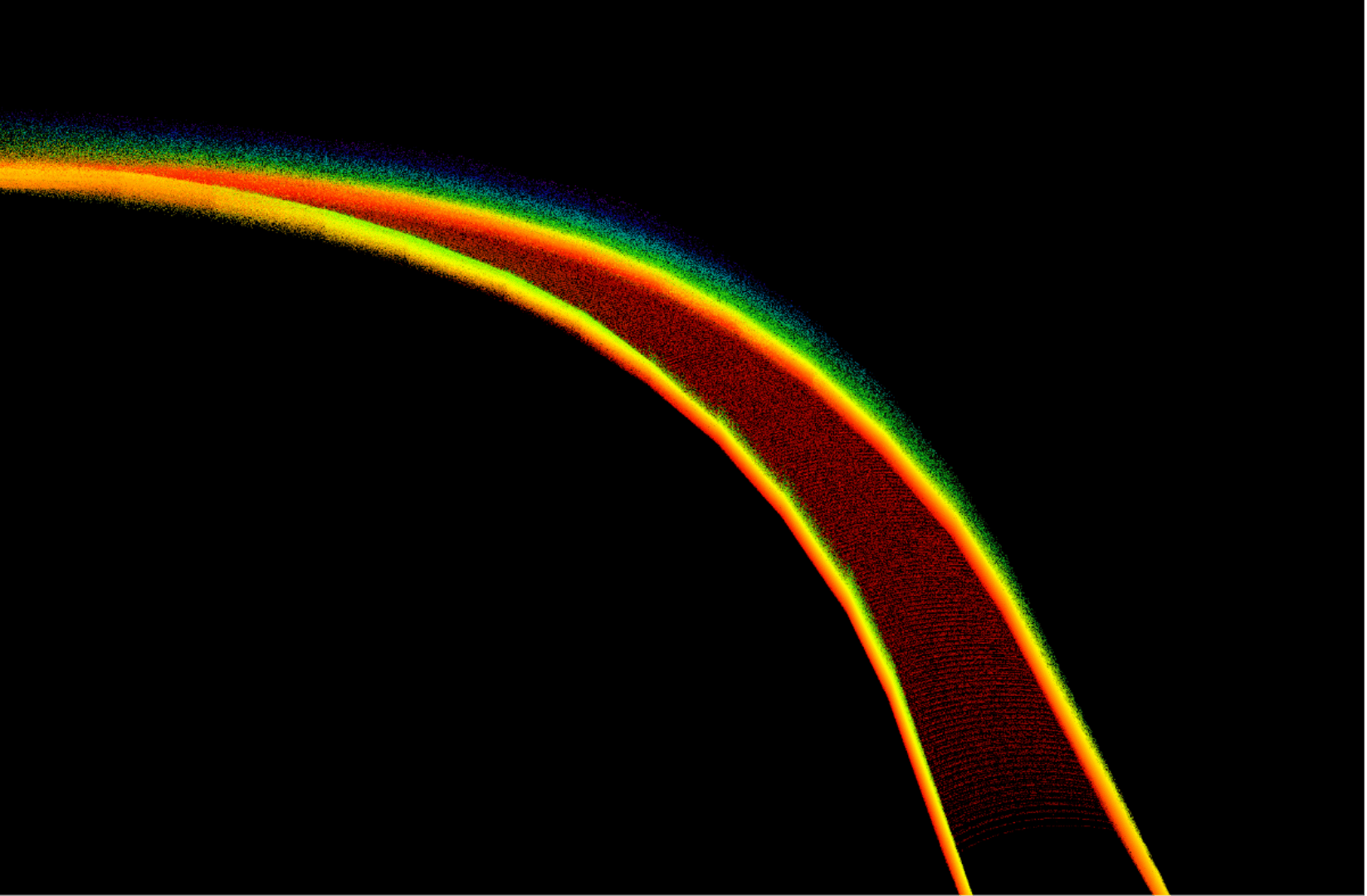} 
		\caption{{Non-Iterative Scan Matching, \textit{Curved Wall}}}
	\end{subfigure}
	\hfill
	\begin{subfigure}[t]{0.3\textwidth}
		\centering
		\includegraphics[width=\linewidth]{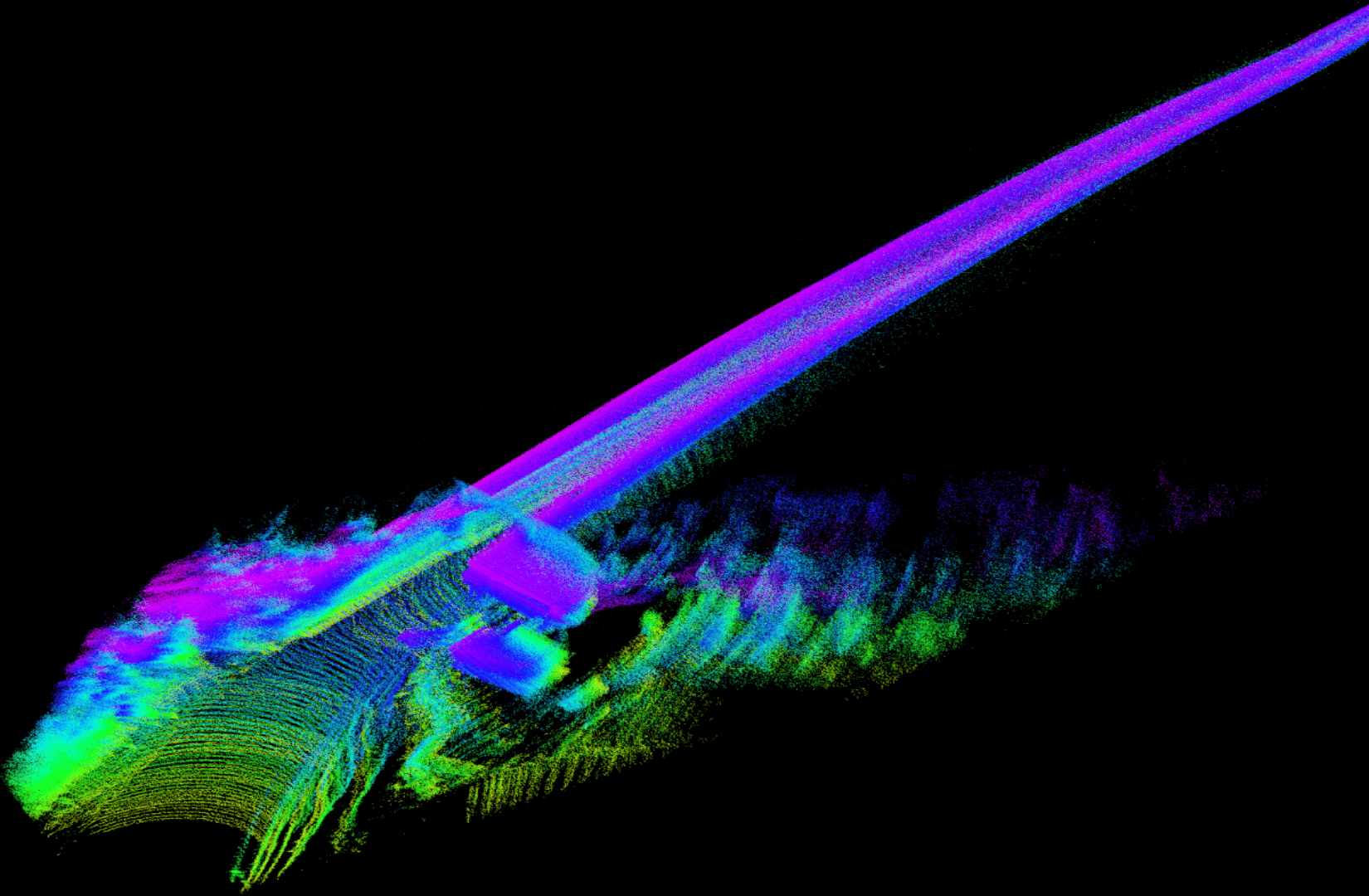}
		\caption{{Non-Iterative Scan Matching, \textit{Baker-Barry Tunnel (Empty)}}}
	\end{subfigure}
	
	\caption{The top row shows results using point-to-point ICP, 
		while the bottom row shows those from the proposed method. 
		Our method alleviates matching ambiguity caused by repetitive geometry.}
	\label{fig:geo_example}
\end{figure*}
\subsection{Experiment Results}
\begin{figure*}[t]
	\centering
	\begin{subfigure}[t]{0.3\textwidth}
		\centering
		\includegraphics[width=\linewidth]{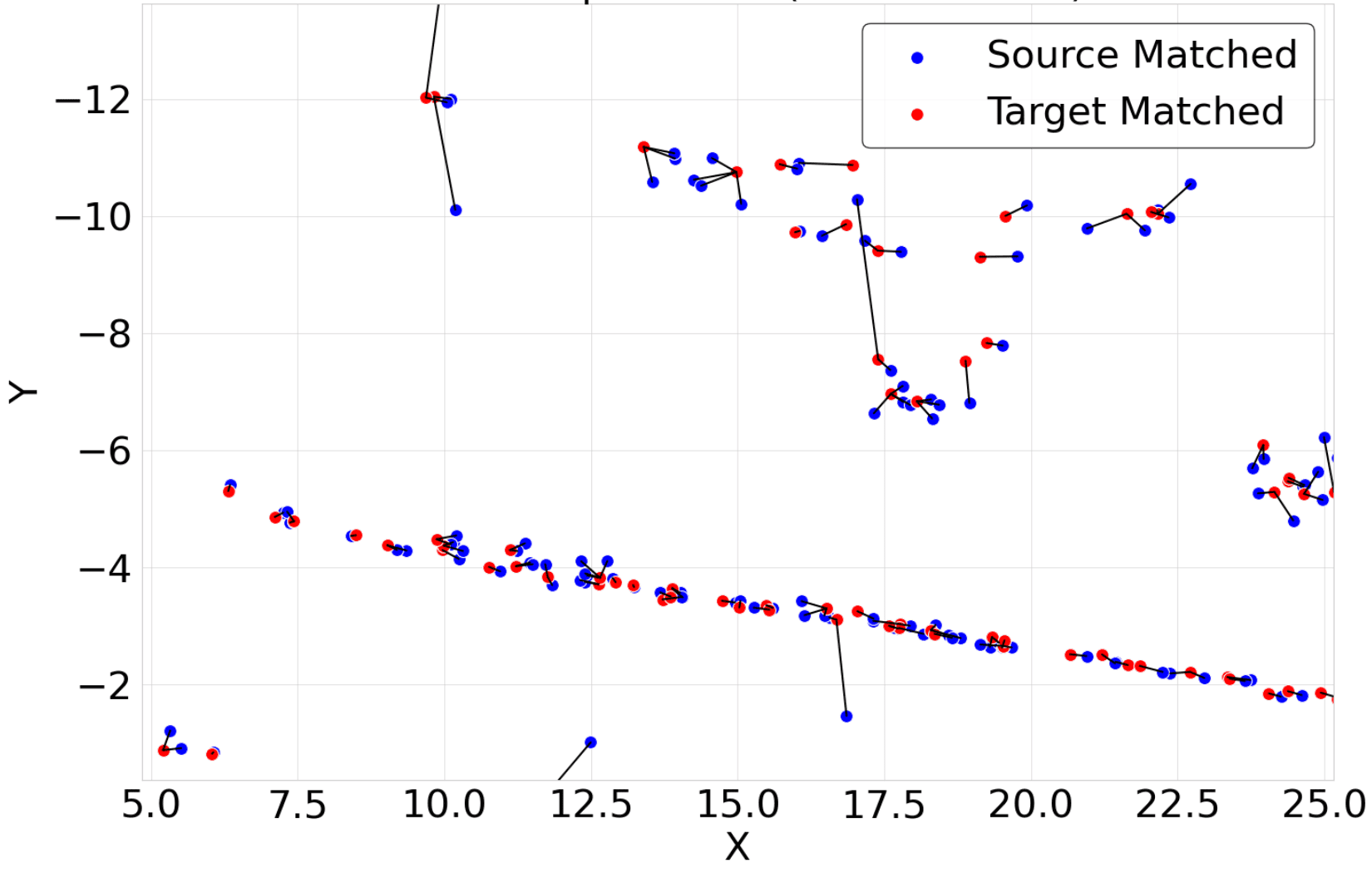}
		\caption{{Sparse, Closest Point Correspondence}}
	\end{subfigure}
	\hfill
	\begin{subfigure}[t]{0.3\textwidth}
		\centering
		\includegraphics[width=\linewidth]{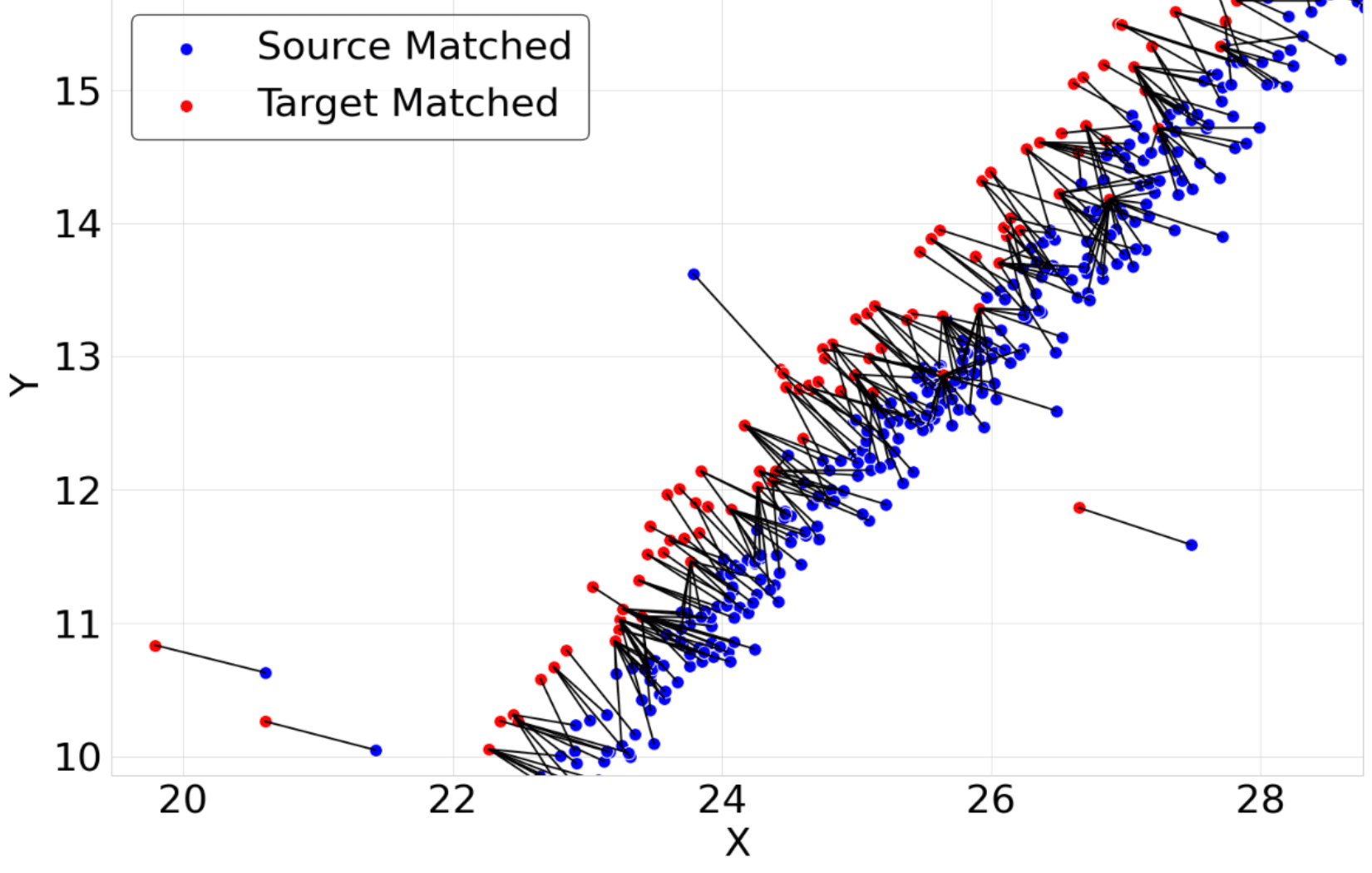}
		\caption{\parbox{\linewidth}{\centering Semi-Dense, Closest Point Correspondence}}

	\end{subfigure}
	\hfill
	\begin{subfigure}[t]{0.3\textwidth}
		\centering
		\includegraphics[width=\linewidth]{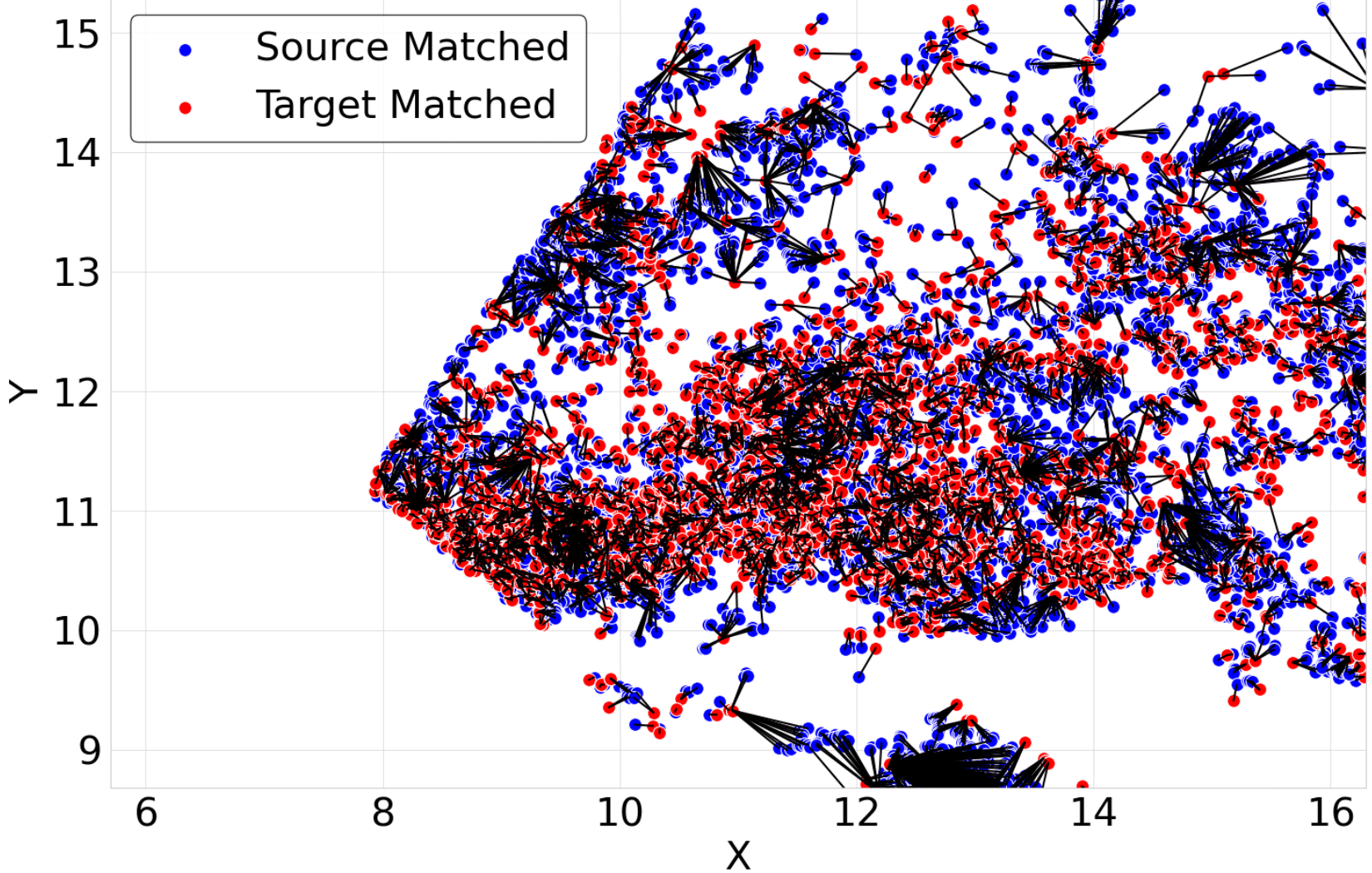}
		\caption{{Dense, Closest Point Correspondence}}
	\end{subfigure}
	
	\vspace{1em} 
	
	\begin{subfigure}[t]{0.3\textwidth}
		\centering
		\includegraphics[width=\linewidth]{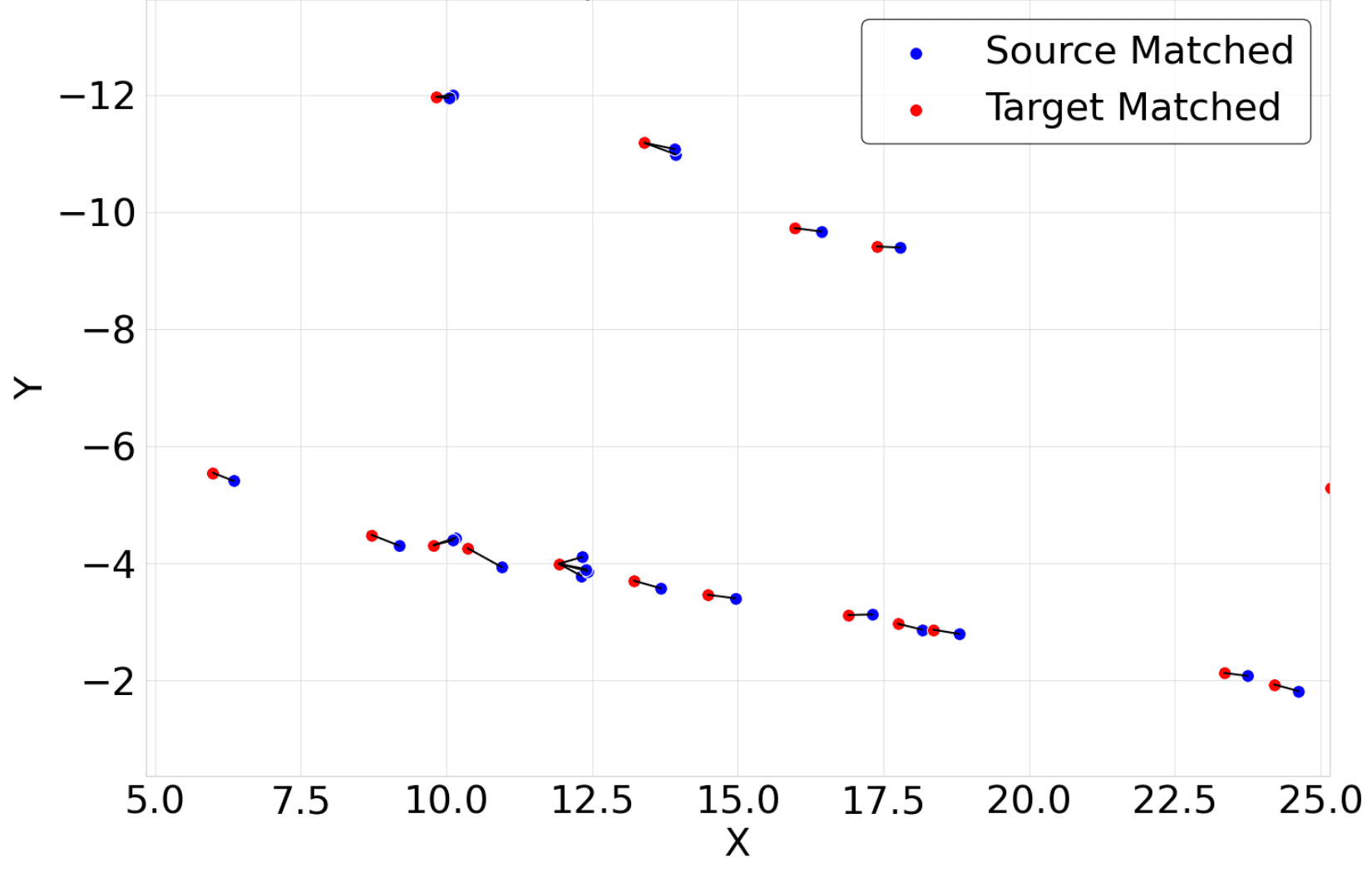}
		\caption{{Sparse, Doppler Correspondence}}
	\end{subfigure}
	\hfill
	\begin{subfigure}[t]{0.3\textwidth}
		\centering
		\includegraphics[width=\linewidth]{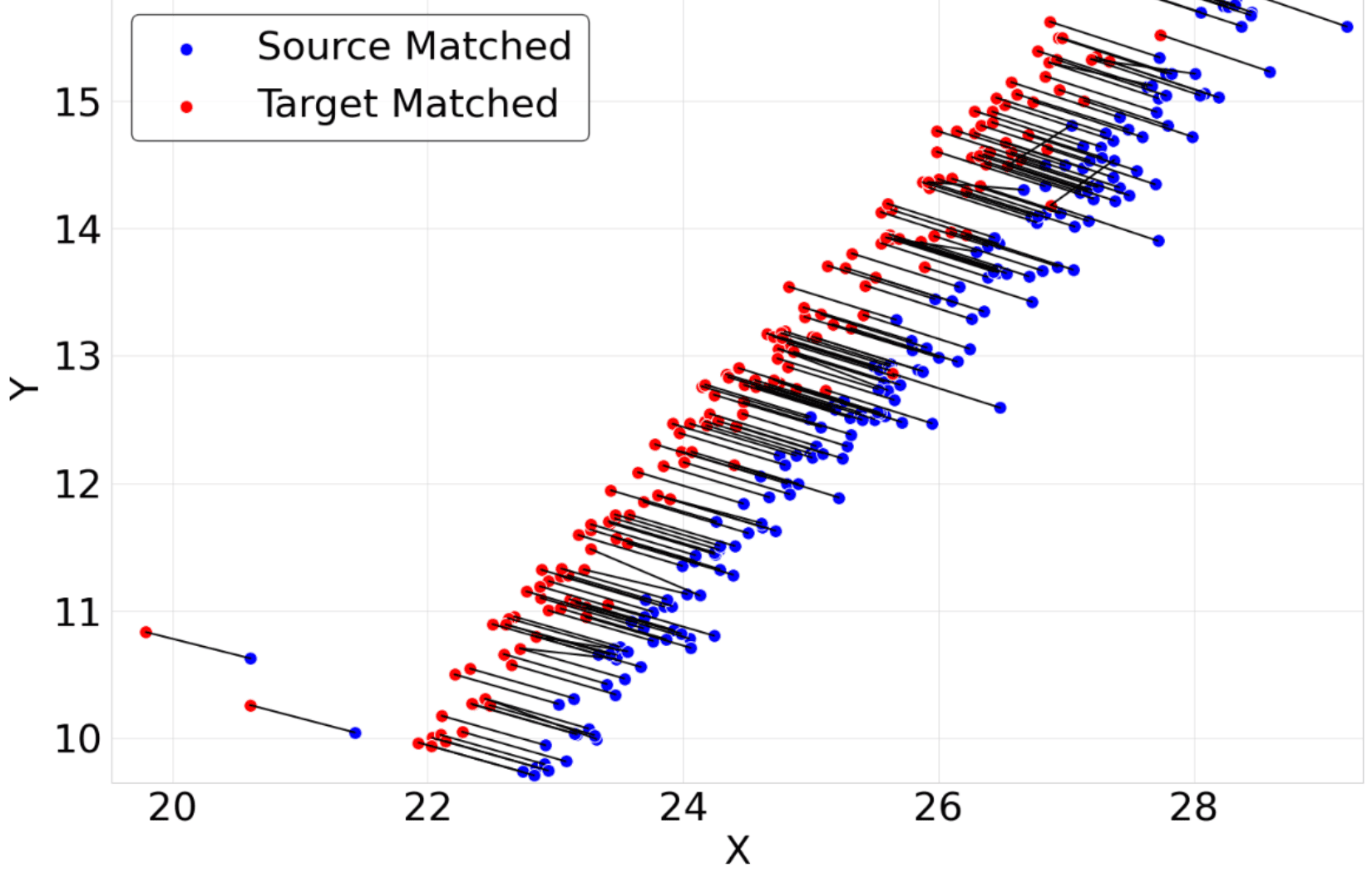}
		\caption{{Semi-Dense, Doppler Correspondence}}
	\end{subfigure}
	\hfill
	\begin{subfigure}[t]{0.3\textwidth}
		\centering
		\includegraphics[width=\linewidth]{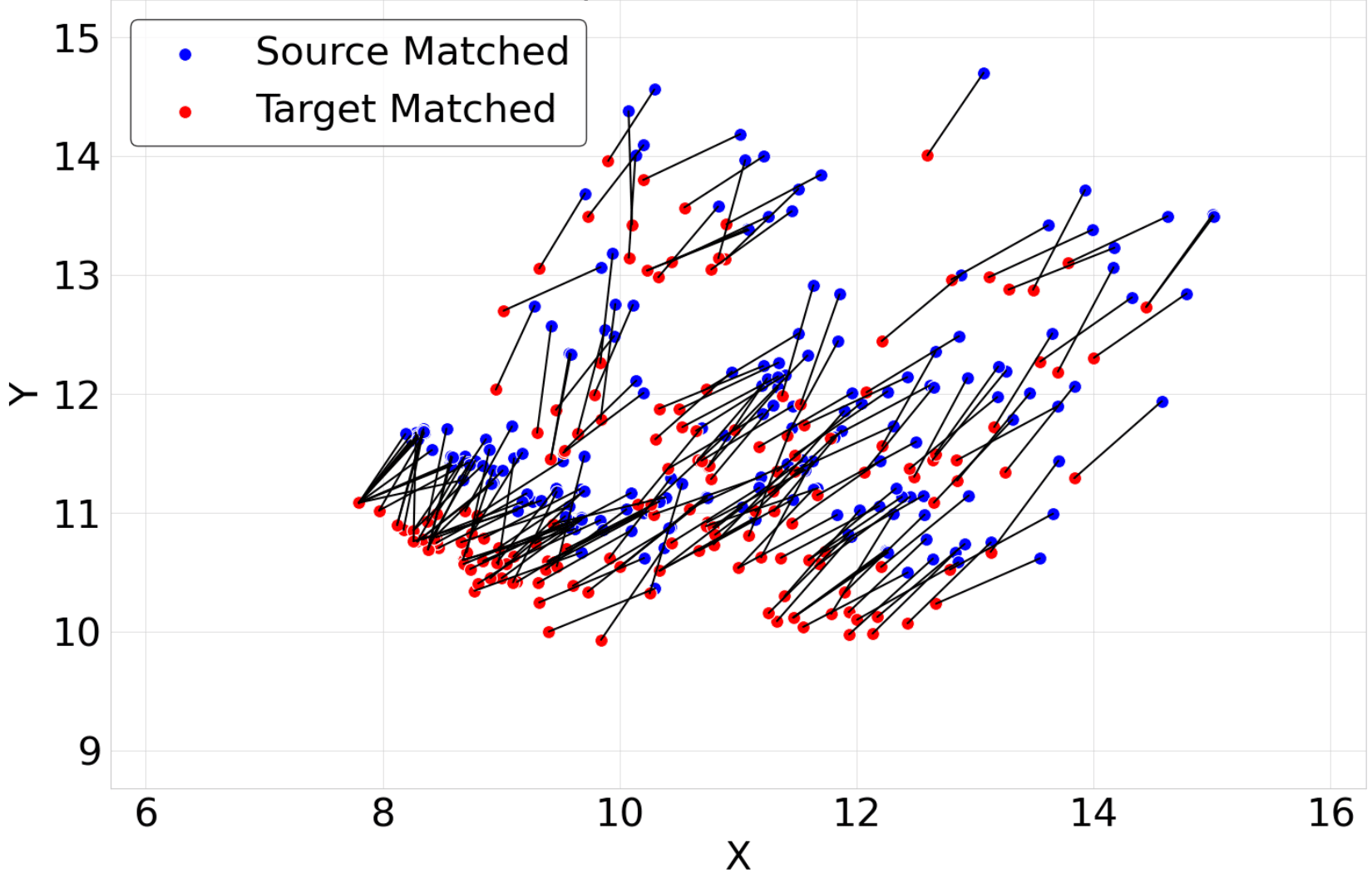}
		\caption{{Dense, Doppler Correspondence}}
	\end{subfigure}
	\caption{Visualization of established correspondences after a single iteration in 
		sparse, semi-dense, and dense point clouds.}
	\label{fig: correspondence}
\end{figure*}
The performance evaluation on sparse 4D radar data compares the proposed method against the standard point-to-point ICP approach, while the open-source implementation of DICP is excluded due to its failure to produce reliable results on sparse 4D radar data. Under these conditions, closest point correspondence, which relies on the spatial overlap of consecutive scans, becomes unreliable~\cite{huang2024less}. By contrast, the proposed correspondence leverages the flow of points~\cite{ding2022self} rather than explicit spatial positions, offering robustness against sparse and noisy data. As shown in Table~\ref{table:results}, Figs.~\ref{fig:trajectory}(a) and (b), and Fig.~\ref{fig: Error_Comparsion}(a), the proposed method attains substantially higher accuracy than that of point-to-point ICP. Moreover, unlike point-to-point ICP, which requires repeated correspondence estimation in 3D space, our approach simplifies the process to a single-step and 1D correspondence estimation.

For semi-dense 4D radar point clouds, the proposed method was tested and compared with DICP. The proposed method's translation errors exhibited a lower median but higher variance than those of DICP, as illustrated in Fig.~\ref{fig: Error_Comparsion}(b). Additionally, rotation errors demonstrated higher median and variance compared with DICP, particularly in sequences with significant rotational motion (\textit{Loop1\_4}). Despite these limitations, the proposed method achieves comparable performance to DICP in other scenarios. Furthermore, it offers computational advantages due to its efficient matching process.

The proposed method is further evaluated on dense 4D LiDAR, particularly in geometrically repetitive environments, and compared with DICP. In static scenarios, \textit{Straight Wall, Curved Wall}, and \textit{ Baker-Barry Tunnel (Empty)}, the proposed method demonstrates comparable performance to DICP while being significantly faster than DICP. Moreover, the method achieves lower rotation error than DICP in \textit{Curved Wall}, possibly due to the Doppler velocity term in DICP having a limited effect on constraining the rotational component during the optimization process. The performance of the proposed method in the \textit{Baker-Barry Tunnel (Vehicles)} is degraded due to a significant number of dynamic objects.

\subsection{Repetitive Geometric Environment} 
In contrast to traditional closest-point correspondence, which relies solely on spatial geometry, the proposed approach leverages the simple functions \( f \) and \( g \) defined in Eq.~\eqref{eq:functions}, which incorporate Doppler velocity. Because these Doppler-based functions do not depend on geometric cues alone, they offer robust performance even when surfaces appear similar or repetitive. As illustrated in Fig.~\ref{fig:geo_example}, this fundamental reliance on velocity information disambiguates correspondences, providing accurate alignment in settings where ICP may fail. While the \textbf{Optimization} process has no difference from other registration techniques, the core difference lies in how correspondences are established (\textbf{Correspondence Estimation}) via Doppler information, rather than through purely closest point matching.

\subsection{Correspondence Visualization}
The matching results of Doppler Correspondence and the closest point correspondence method are visualized on sparse, semi-dense, and dense point clouds. The correspondences established after the single iteration are performed, as shown in Fig.~\ref{fig: correspondence}. With only a single iteration, Doppler Correspondence achieves more coherent and consistent matches than the closest point approach across all levels of point cloud density. 

In the sparse 4D radar point clouds, Doppler Correspondence generates significantly fewer correspondences than a number of points, as illustrated in Figs.~\ref{fig: correspondence}(a) and (d). However, those correspondences are high quality, leading to better alignment performance compared with the closest-point method, as shown in Fig.~\ref{fig: Error_Comparsion}(a). This result aligns with the \textit{less is more} principle discussed in~\cite{huang2024less}, where using fewer but more reliable correspondences can be advantageous in highly noisy data. For semi-dense 4D radar point cloud, while the closest point correspondence often fails to establish reliable matches, Doppler Correspondence succeeds in maintaining robust and consistent pairings, even when only a single iteration is performed, as illustrated in Figs.~\ref{fig: correspondence}(b) and (e). For dense point clouds with repetitive geometric structures (\textit{Baker-Barry Tunnel (Empty)}), Doppler Correspondence can produce clear and stable matches by effectively mitigating the ambiguity that may degrade the performance of ICP approaches, as depicted in Figs.~\ref{fig: correspondence}(c) and (f). 

\section{Limitations}
\label{sec:Limitation-Condition}

While Doppler Correspondence, introduced in Section~\ref{sec: Methodology}, offers notable advantages in terms of computational efficiency and robustness in repetitive geometric environments, it also comes with certain limitations. Specifically, the key scenarios where the method may fail are highlighted below. We also clarify the main assumptions underlying Eqs.~\eqref{eq:dot_product_p} and \eqref{eq:dot_product_q}. Furthermore, potential scenarios where these assumptions may fail in practice are discussed.
\subsection{Key Assumptions and Their Breakdown}
For identification of the case where Eqs.~\eqref{eq:dot_product_p} and ~\eqref{eq:dot_product_q} hold, the motion of a point \(p_i\) can be considered as a combination of translation \(\Delta t \in \mathbb{R}^{3 \times 1}\) and rotation \(\Delta R \in SO(3)\) from the sensor between a time interval \(\Delta T\). Suppose \(p_i(T)\) is the point \(p_i\) at time \(T\). After the time interval \(\Delta T\), the point moves according to:
\begin{equation}
	\label{assumption_1}
	p_i\bigl(T + \Delta T\bigr)
	\;=\;
	\Delta R \, p_i(T) \;+\; \Delta t,
\end{equation}
where \(\Delta R = \exp(\Delta T\, \omega^\times)\). Here, \(\omega \in \mathbb{R}^3\) is the angular velocity, and \(\omega^\times\) denotes its skew-symmetric matrix. 

Assuming \(\Delta T \; \|\omega\|_2\) \(\ll\) 1, \(\Delta R\) can be linearized as:
\begin{equation}
    \label{small_angle_approx}
    \Delta R \;\approx\; I + \Delta T\, \omega^\times.
\end{equation}

Substituting this into Eq.~\eqref{assumption_1} gives
\[
	p_i\bigl(T + \Delta T\bigr) \;-\; p_i(T)
\;=\;
(\Delta T\, \omega^\times)\, p_i(T)
\;+\;
\Delta t.
\]
Dividing both terms by \(\Delta T\) gives:
\begin{equation}
	\label{equ_1}
	\frac{p_i\bigl(T + \Delta T\bigr) \;-\; p_i(T)}{\Delta T} = w^\times p_i(T) + \rho,
\end{equation}
where \(\rho = \frac{\Delta t}{\Delta T}\). Assuming constant \(\omega\) and \(\rho\) during a short consecutive scan, as \(\Delta T \rightarrow 0\), then:
\begin{equation}
    \label{constant_assump}
    \dot{p}_i(T)=w^\times p_i(T) + \rho.
\end{equation}

Taking the inner product of both sides with the unit vector 
\[
n_{p_{i}}(T) 
\;=\; 
\frac{p_i(T)}{\|p_i(T)\|_2},
\]
we obtain
\begin{equation}
	\label{eq2}
	v_{p_i}(T)= n_{p_{i}}(T) \cdot \rho, 
\end{equation}
\[
(\because n_{p_{i}}(T)  \cdot (\omega \times p_i(T)) = 0).
\]
Therefore, for any \(T\) between a short \(\Delta T\), \(\omega\) and \(\rho\) do not significantly change, and Eq.~\eqref{eq2} can be used.

In our framework, if we set \(p_i \in \mathcal{P}\) and \(q_j \in \mathcal{Q}\) as true correspondences from a small time interval \(\Delta T\), Eq.~\eqref{equ_1} can be written as:
\begin{equation}
	\label{eq4}
		\frac{q_j \;-\; p_i}{\Delta T} = w^\times p_i + \rho.
\end{equation}
Applying Eq.~\eqref{eq2} at the points where \(p_i\) and \(q_j\) are measured:
\begin{equation}
	v_{p,i}= n_{p_{i}} \cdot \rho, 
\end{equation}
\begin{equation}
v_{q,j}= n_{q_{j}}\cdot \rho.
\end{equation}
Taking inner product of both sides \(n_{p,i}\) to Eq.~\eqref{eq4}, Eq.~\eqref{eq:radial_p} can be derived:
\[
\frac{q_j \;-\; p_i}{\Delta T}\cdot n_{p_i} = (w^\times p_i + \rho) \cdot n_{p_i} = \rho \cdot n_{p_i} = v_{p_i}.
\]
Multiplying both sides by $\Delta T$ and $\|p_i\|_2$, we obtain Eq.~\eqref{eq:dot_product_p}:
\[
(q_j - p_i)\cdot p_i = v_{p,i} \, \Delta T \, \|p_i\|_2.
\]

Derivation of Eq.~\eqref{eq:dot_product_q} starts from the relationship \(q_j = \Delta R p_i~+~\Delta t\), which can be expressed as:
\[
\Delta R^T q_j - \Delta R^T \Delta t = p_i.
\]
Thus, \(p_i - q_j\) can be formulated as:
\[
	p_i - q_j = (I - \Delta T \omega^\times)q_j - (I - \Delta T \omega^\times)\Delta t - q_j
\]
\[
= -\Delta T \omega^\times q_j - (I - \Delta T \omega^\times)\Delta t.
\]
Divide the above with \(\Delta T\):
\[
\frac{p_i - q_j }{\Delta T} = -\omega^\times q_j - (I-\Delta T \omega^\times) \rho,
\]
and take inner product with \(n_{q_j}\):
\[
\frac{p_i - q_j }{\Delta T}\cdot n_{q_j} = - (I-\Delta T \omega^\times) \rho \cdot n_{q_j}
\]
\[
= -\rho \cdot n_{q_j} + (\Delta T \omega^\times \rho)\cdot n_{q_j}
\]
\[
= -v_{q,j}+ (\Delta T \omega^\times \rho)\cdot n_{q_j}.
\]
Multiplying both sides by $\Delta T$ and $\|q_j\|_2$, we obtain:
\[
(q_j - p_i)\cdot q_j = \underbrace{\Delta T~v_{q_j}\|q_j\|_2}_{\text{Doppler velocity term}} - \underbrace{\Delta T^2(\omega^\times~\rho)\cdot q_j}_{\text{rotation term}}.
\]
Since $\Delta T^2 \approx 0$, the \textit{rotation term} can be neglected, leading to the Eq.~\eqref{eq:dot_product_q}:
\[
(q_j - p_i)\cdot q_j \approx v_{q_j}~\Delta T~\|q_j\|_2.
\]
Therefore, the \textit{sufficient conditions} required to derive Eqs.~\eqref{eq:dot_product_p} and~\eqref{eq:dot_product_q} are that 
\textbf{$\boldsymbol{\Delta T \|\omega\|_2 \ll 1}$}, as \textit{assumed in} Eq.~\eqref{small_angle_approx}, 
and that both $\boldsymbol{\omega}$ and $\boldsymbol{\rho}$ remain \textbf{constant over a short consecutive scan}, 
as \textit{assumed in} Eq.~\eqref{constant_assump}. 

The first condition, $\Delta T \|\omega\|_2 \ll 1$, is necessary to linearize the rotation matrix $\Delta R = \exp(\Delta T\,\omega^\times)$. This simplification enables the relative motion between $p_i$ and $q_j$ to be approximated using linear vector operations, which is essential for establishing a direct relationship between spatial displacement and Doppler velocity. 

The second condition, assuming that both $\omega$ and $\rho$ remain constant during a short interval $\Delta T$, is used in Eq.~\eqref{constant_assump}. This implies that the velocity of a point $p_i$ can be written as a time-invariant expression, ensuring that Doppler velocity measurements are consistent across a short time window. If $\omega$ or $\rho$ varies significantly, the radial velocity observed from Doppler measurements would also change, making it difficult to formulate a consistent correspondence constraint between $p_i$ and $q_j$.

Violating this condition may lead to degradation or failure of the proposed method. The following examples illustrate the cases where the assumption holds and it does not:
\subsubsection{Gentle Turn (Assumption Holds)}
\label{sec:GentleTurn}
The scenario where the assumption remains valid is illustrated in Fig.~\ref{fig:combined_subfigures}(a). The vehicle travels at a moderate speed 
\(\bigl(v \approx 8.3\,\mathrm{m/s}\bigr)\) with a small angular velocity \(\bigl(\|\omega\|_2 \approx 0.1\,\mathrm{rad/s}\bigr)\), 
and the sensor operates at \(\Delta T = 0.083\,\mathrm{s}\). In this case,
\[
\Delta T\,\|\omega\|_2 
\;=\; 0.083 \times 0.1 
\;=\; 0.0083 
\;\ll\; 1.
\]
In this scenario, both Eqs.~\eqref{eq:dot_product_p} and \eqref{eq:dot_product_q} hold, resulting in accurate Doppler-based odometry with minimal drift.
\subsubsection{U-Turn (Assumption Fails)}
\label{sec:UTurn}
\begin{figure}[t!]
	\centering
	\begin{subfigure}[b]{\linewidth}
		\centering
		\includegraphics[width=0.95\linewidth]{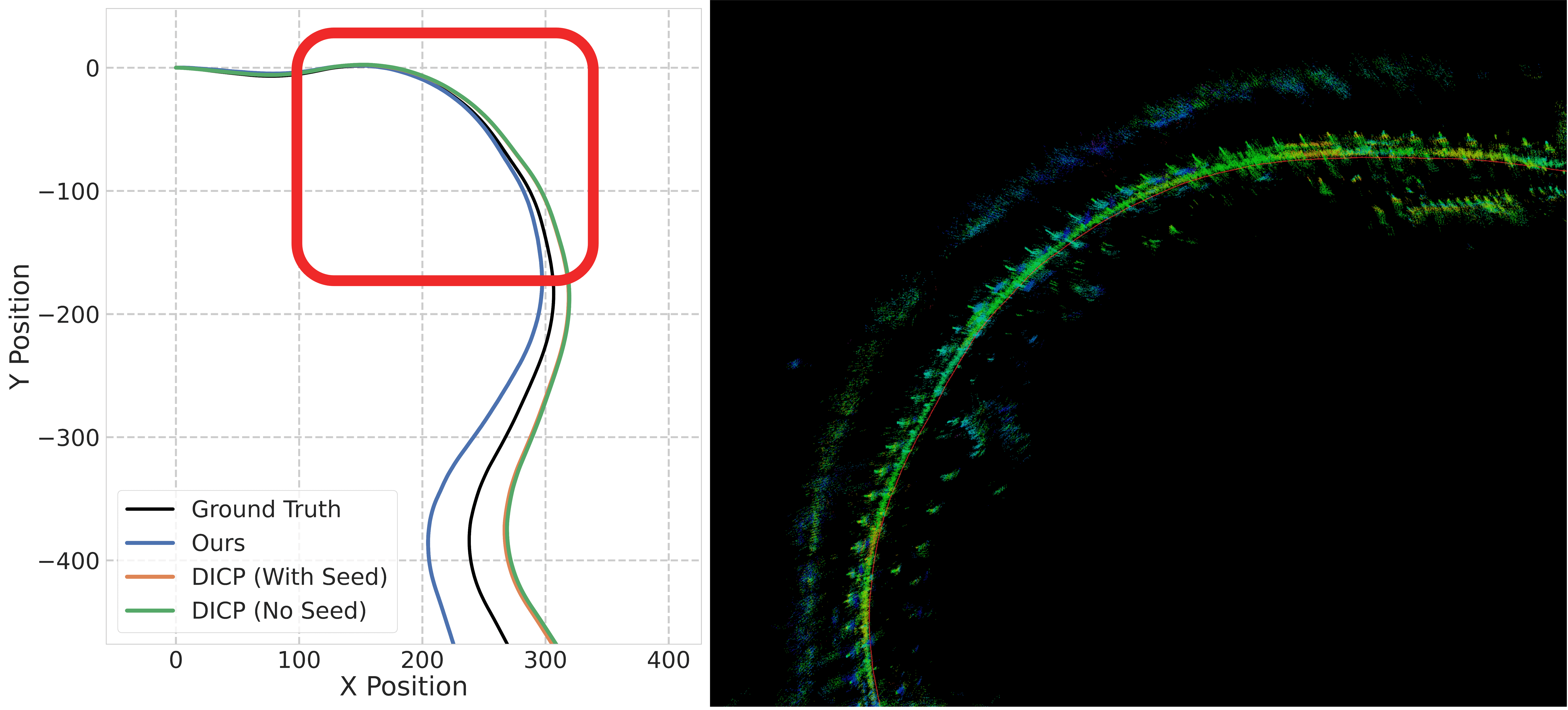}
		\caption{Gentle Turn}
		\label{fig:subfig1}
	\end{subfigure}
	\begin{subfigure}[b]{\linewidth}
		\centering
		\includegraphics[width=0.95\linewidth]{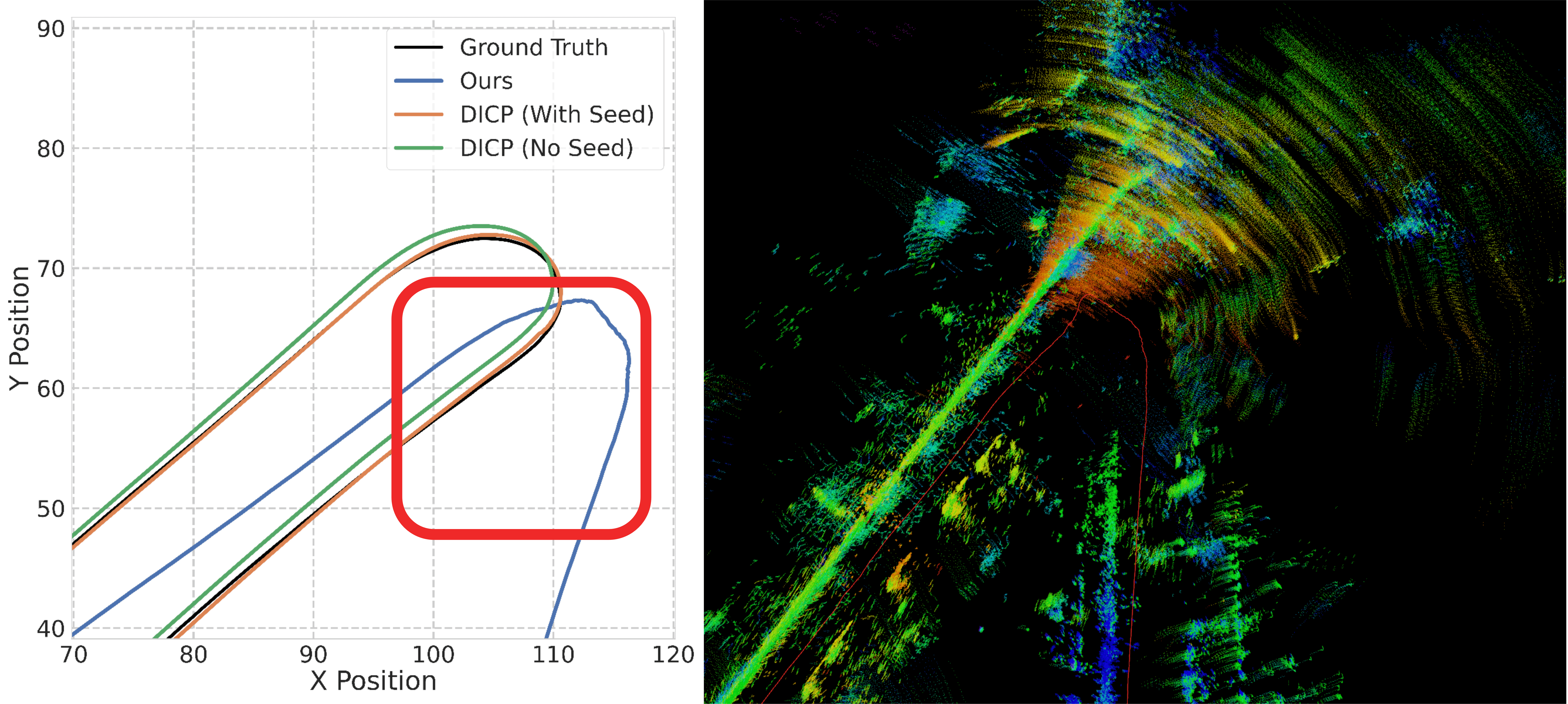}
		\caption{U Turn}
		\label{fig:subfig2}
	\end{subfigure}
	\caption{(a) corresponds to the gentle turn example from \textit{Loop1\_2}, and (b) corresponds to the U-turn example from \textit{Loop1\_4}. Each left figure shows the trajectories produced by each method, and the right figure illustrates the map constructed in the red area highlighted in the left figure.}
	\label{fig:combined_subfigures}
\end{figure}

By contrast, Fig.~\ref{fig:combined_subfigures}(b) shows a scenario where the assumption is invalid. The angular velocity of the vehicle is significantly higher than that of \textit{Gentle Turn} scenario
\(\bigl(\|\omega\|_2 \approx 0.56\,\mathrm{rad/s}\bigr)\). 
Even though \(\Delta T\) remains \(0.083\,\mathrm{s}\), the product
\[
\Delta T\,\|\omega\|_2 
\;=\; 0.083 \times 0.56 
\;=\; 0.04648,
\]
is no longer “small.” As a result, the odometry output can deviate significantly when sharp U-turns occur.
\begin{algorithm}[t]
	\caption{Doppler Correspondence-Based Iterative Closest Point for 4D Ranging Sensors (Doppler 4D-ICP)}
	\begin{algorithmic}
		\Require 
		$\mathcal{P}$ (source point cloud), 
		$\mathcal{Q}$ (target point cloud), 
		$\tau_{\text{spatial}}$ (spatial distance threshold),
		$\tau_{\text{doppler}}$ (doppler distance threshold)
		\Ensure 
		$\hat{R}, \hat{t}$
		\State Compute $f(\mathcal{P}), g(\mathcal{Q})$
		\State $\mathcal{Q'} \gets \texttt{FindClosest}(f(\mathcal{P}),g(\mathcal{Q}))$
		\State \textbf{Masking Outliers:}
		\State \quad $\mathcal{M} \gets \|\mathcal{P} - \mathcal{Q'}\|_2 \leq \tau_{\text{spatial}} \text{ and } \left| f(\mathcal{P}) - g(\mathcal{Q'}) \right| \leq \tau_{\text{doppler}}$
		\State \quad $\mathcal{C}_{\text{doppler}} \gets \mathcal{P}[\mathcal{M}], \mathcal{Q}^{\prime}[\mathcal{M}]$
		\While{not converged}
		\State $\mathcal{Q'} \gets \texttt{FindClosest}(\mathcal{P}, \mathcal{Q})$
		\State \textbf{Masking Outliers:}
		\State \quad $\mathcal{M} \gets \|\mathcal{P} - \mathcal{Q}^{\prime}\|_2 \leq \tau_{\text{spatial}}$
		\State \quad $\mathcal{C}_{\text{spatial}} \gets \mathcal{P}[\mathcal{M}], \mathcal{Q}^{\prime}[\mathcal{M}]$
		\State \textbf{Predict Transformation:}
		\State \quad $\delta{R}, \delta{t} \gets \underset{R, t}{\operatorname{arg\,min}} \;
		(1 - \alpha) \sum_{(p, q) \in \mathcal{C}_{\text{spatial}}} \rho(\| q - (R p + t) \|_2) 
		+ \alpha \sum_{(p, q) \in \mathcal{C}_{\text{doppler}}} \rho(\| q - (R p + t) \|_2)$
		\State \textbf{Update:}
		\State \quad $\mathcal{P} \gets \texttt{TransformSourcePoint}(\delta R, \delta t, \mathcal{P})$
		\State \quad $\hat{R},\hat{t}\gets \texttt{UpdateTransformation}(\delta{R},\delta{t},\hat{R},\hat{t})$
		\EndWhile
		
		\Return $\hat{R}, \hat{t}$
	\end{algorithmic}
	\label{wicp}
\end{algorithm}

\section{Practical Use of Doppler Correspondence}
\label{practical usage}
Doppler Correspondence may offer versatile applications due to its simplicity. A straightforward implementation is its integration with ICP in a weighted form, referred to as Doppler 4D-ICP, as described in Algorithm~\ref{wicp}. The process begins by computing \(f(\mathcal{P})\) and \(g(\mathcal{Q})\) from the point clouds and establishing Doppler Correspondence. In this setup, correspondences estimated from Doppler Correspondence remain fixed throughout the ICP iterations, while closest point correspondences are dynamically updated. This combination enables the optimization process to leverage two distinct correspondence characteristics: spatial information and Doppler information. To further enhance robustness against outliers and noise, we additionally incorporate a Huber loss function \( \rho(\cdot) \) as a robust kernel during optimization.

To evaluate the effectiveness of the proposed approach, experiments are conducted using a semi-dense 4D radar dataset, \textit{Loop 1}, and dense 4D LiDAR datasets, \textit{Baker-Barry Tunnel (Empty, Vehicles)}. The parameters are configured as follows: for the semi-dense 4D radar dataset, $\tau_{\text{spatial}} = 3$, $\tau_{\text{doppler}} = 10$, and $\alpha = 0.7$. For the dense 4D LiDAR datasets, the parameters are set to $\tau_{\text{spatial}} = 2$, $\tau_{\text{doppler}} = 5$, and $\alpha = 0.9$. 

The experimental results are presented on the semi-dense 4D radar dataset in Table~\ref{tab:semi_dense_wicp} and the dense 4D LiDAR dataset in Table~\ref{tab:dense_wicp}.  Notably, in repetitive geometric environments, geometric ambiguity is difficult to resolve without incorporating Doppler information, leading to divergence in conventional scan-matching methods. By leveraging Doppler Correspondence, the proposed approach effectively fuses spatial and Doppler information, 
leading to more reliable and consistent trajectory estimation.

\section{Conclusion} 
Overall, the results indicate that Doppler Correspondence does not always guarantee better performance compared with the closest point method. Scenarios with high angular velocity may cause the violation of the aforementioned assumption and degradation of odometry performance. However, the proposed method offers distinct advantages, such as its weak dependence on the environment's geometry, which sets it apart from closest point correspondence. These features enable robust and efficient correspondence estimation, making it a valuable complement to traditional methods in specific applications. Furthermore, its simplicity makes it easy to use and integrate with other methods. While further improvements are needed for broader applicability, the proposed method holds novelty as it leverages Doppler information in correspondence for the first time.
\label{sec: con}

\begin{table}[t]
    \centering
    \small
    \caption{Odometry performance on the semi-dense 4D radar dataset.  
    Each cell reports the absolute pose errors~\cite{grupp2017evo} (in meters).  
    Best results are in bold, second-best are underlined.}
    \label{tab:semi_dense_wicp}
    \resizebox{\linewidth}{!}{%
    \begin{tabular}{c|cccc}
        \hline
        \multirow{2}{*}{\textbf{Method}} & \multicolumn{4}{c}{\textbf{Sequence}} \\
        \cline{2-5}
        & \textit{loop1\_0} & \textit{loop1\_1} & \textit{loop1\_3} & \textit{loop1\_4} \\ \hline
        ICP point-to-point               & 61.72 & \underline{38.71} & 146.65 & 3.15 \\
        ICP point-to-plane               & 55.18 & 43.40 & 127.52 & \underline{2.89} \\
        Generalized-ICP~\cite{segal2009generalized} & 64.21 & 57.28 & 138.41 & 3.30 \\
        VGICP~\cite{koide2021voxelized}  & \textbf{50.86} & 61.86 & 141.37 & 4.09 \\
        DICP~\cite{hexsel2022dicp}       & 60.15 & 68.60 & 143.24 & 2.94 \\
        Doppler Correspondence (Ours)    & \underline{54.87} & 50.04 & \textbf{101.74} & 7.49 \\
        Doppler 4D-ICP (Ours)              & 63.25 & \textbf{37.74} & \underline{114.76} & \textbf{2.80} \\ \hline
    \end{tabular}
    }
\end{table}
\begin{table}[t]
    \centering
    \small
    \caption{Odometry performance on the dense 4D LiDAR dataset.  
    Each cell reports the absolute pose errors~\cite{grupp2017evo} (in meters).  
    Best results are in bold, second-best are underlined. Note that Div. indicates the trajectory of the algorithm diverged.}
    \label{tab:dense_wicp}
    \resizebox{\linewidth}{!}{%
    \begin{tabular}{c|clcl}
        \hline
        \multirow{2}{*}{\textbf{Method}} & \multicolumn{4}{c}{\textbf{Sequence}} \\ 
        \cline{2-5}
        & \multicolumn{2}{c}{\textit{\begin{tabular}[c]{@{}c@{}}Baker-Barry Tunnel \\ (Empty)\end{tabular}}} & \multicolumn{2}{c}{\textit{\begin{tabular}[c]{@{}c@{}}Baker-Barry Tunnel \\ (Vehicles)\end{tabular}}} \\
        \hline
        ICP point-to-point               & \multicolumn{2}{c}{Div.} & \multicolumn{2}{c}{Div.} \\
        ICP point-to-plane               & \multicolumn{2}{c}{Div.} & \multicolumn{2}{c}{Div.} \\
        Generalized-ICP~\cite{segal2009generalized} & \multicolumn{2}{c}{Div.} & \multicolumn{2}{c}{Div.} \\
        VGICP~\cite{koide2021voxelized} & \multicolumn{2}{c}{Div.} & \multicolumn{2}{c}{Div.} \\
        DICP~\cite{hexsel2022dicp} & \multicolumn{2}{c}{65.27} & \multicolumn{2}{c}{\underline{37.12}} \\
        Doppler Correspondence (Ours)    & \multicolumn{2}{c}{\underline{20.16}} & \multicolumn{2}{c}{47.54} \\
        Doppler 4D-ICP (Ours)              & \multicolumn{2}{c}{\textbf{14.50}} & \multicolumn{2}{c}{\textbf{24.06}} \\
        \hline
    \end{tabular}
    }
\end{table}
\vspace{-1mm}
\section*{Acknowledgments}
This work was supported by the National Research Foundation of Korea (NRF) grants funded by the MSIT (No. 2023R1A2C2003130) and the Ministry of Education (No. 2020R1A6A1A03040570).

\label{sec: con}
\bibliographystyle{plainnat}
\bibliography{references}

\end{document}